\pdfoutput=1
\PassOptionsToPackage{dvipsnames}{xcolor} 
\documentclass{fairmeta}

\usepackage[utf8]{inputenc}

\usepackage{amsmath}
\usepackage{amssymb}
\usepackage{amsfonts}
\usepackage{mathtools}
\usepackage{bbm}
\usepackage{mathrsfs}
\usepackage{nicefrac}

\usepackage{float}
\usepackage{placeins} 
\usepackage{wrapfig}
\usepackage{algorithm}
\usepackage[noend]{algpseudocode}

\usepackage{tikz}
\usetikzlibrary{positioning, arrows.meta, calc, decorations.pathreplacing, shapes.geometric}

\usepackage{enumitem}
\usepackage{pifont}
\usepackage{xspace}
\usepackage{comment} 

\usepackage{palatino}
\newif\ifmultifactor\multifactortrue
\ifdefined\SDMnomultifactor\multifactorfalse\fi
\newif\ifneuripsversion

\renewcommand{\ast}{\star}

\renewcommand{\epsilon}{\varepsilon}

\newcommand{\gb}{\mathbf{g}}

\def\ddefloop#1{\ifx\ddefloop#1\else\ddef{#1}\expandafter\ddefloop\fi}
\def\ddef#1{\expandafter\def\csname bb#1\endcsname{\ensuremath{\mathbb{#1}}}}
\ddefloop ABCDEFGHIJKLMNOPQRSTUVWXYZ\ddefloop
\def\ddefloop#1{\ifx\ddefloop#1\else\ddef{#1}\expandafter\ddefloop\fi}
\def\ddef#1{\expandafter\def\csname b#1\endcsname{\ensuremath{\mathbf{#1}}}}
\ddefloop ABCDEFGHIJKLMNOPQRSTUVWXYZ\ddefloop
\def\ddef#1{\expandafter\def\csname sf#1\endcsname{\ensuremath{\mathsf{#1}}}}
\ddefloop ABCDEFGHIJKLMNOPQRSTUVWXYZ\ddefloop
\def\ddef#1{\expandafter\def\csname c#1\endcsname{\ensuremath{\mathcal{#1}}}}
\ddefloop ABCDEFGHIJKLMNOPQRSTUVWXYZ\ddefloop
\def\ddef#1{\expandafter\def\csname h#1\endcsname{\ensuremath{\widehat{#1}}}}
\ddefloop ABCDEFGHIJKLMNOPQRSTUVWXYZ\ddefloop
\def\ddef#1{\expandafter\def\csname hc#1\endcsname{\ensuremath{\widehat{\mathcal{#1}}}}}
\ddefloop ABCDEFGHIJKLMNOPQRSTUVWXYZ\ddefloop
\def\ddef#1{\expandafter\def\csname t#1\endcsname{\ensuremath{\widetilde{#1}}}}
\ddefloop ABCDEFGHIJKLMNOPQRSTUVWXYZ\ddefloop
\def\ddef#1{\expandafter\def\csname tc#1\endcsname{\ensuremath{\widetilde{\mathcal{#1}}}}}
\ddefloop ABCDEFGHIJKLMNOPQRSTUVWXYZ\ddefloop
\def\ddefloop#1{\ifx\ddefloop#1\else\ddef{#1}\expandafter\ddefloop\fi}
\def\ddef#1{\expandafter\def\csname scr#1\endcsname{\ensuremath{\mathscr{#1}}}}
\ddefloop ABCDEFGHIJKLMNOPQRSTUVWXYZ\ddefloop

\newcommand{\sz}{\phantom{1}}

\newcommand{\subr}[1]{\raisebox{-0.5ex}{\scriptsize\textcolor{red}{\!\phantom{+}#1}}}
\newcommand{\subg}[1]{\raisebox{-0.5ex}{\scriptsize\textcolor{ForestGreen}{\!\phantom{-}#1}}}

\newcommand{\mysp}{\phantom{0}}
\newcommand{\evalcite}[1]{\citeyearpar{#1}}
\newcommand{\SDMfitwidetable}[1]{#1}

\title{Sparse Delta Memory: Scaling the State of Linear RNNs through Sparsity}

\author{Loïc Cabannes$^{1, 2}$}
\author{Pierre-Emmanuel Mazar\'e$^1$}
\author{Ilze Amanda Auzina$^{1, 3}$}
\ifmultifactor\author{Th\'eo Le Pendeven$^1$}\fi
\author{Gergely Szilvasy$^1$}
\author{Matthijs Douze$^1$}
\author{Maria Lomeli$^1$}
\author{Justin Carpentier$^2$}
\author{Gabriel Synnaeve$^1$}
\author{Herv\'e J\'egou$^1$}

\affiliation{$^1$Meta FAIR}
\affiliation{$^2$Inria Paris \& ENS-PSL University}
\affiliation{$^3$University of Tübingen}

\date{\today}

\abstract{\ifmultifactor\vspace*{-0.30cm}\fi
Linear attention models allow a fixed state size and a fixed amount of compute per token. However, due to their limited state size, linear attention models fall behind in long-context recall compared to softmax-attention-based transformer architectures. Increasing the state size of linear attention improves recall performance but at the cost of higher FLOPs. In this work, we introduce \textbf{Sparse Delta Memory\,(SDM)}, an architecture that scales the hidden state of gated linear RNNs to orders of magnitude higher capacity using a sparse addressing scheme. SDM extends the Gated DeltaNet architecture by replacing the dense key-value outer product with sparse reads and writes to a large explicit memory.
We show that, under an isoFLOP constraint and with an identical number of parameters, a higher state memory capacity significantly improves performance on in-context learning and long-context retrieval tasks. Moreover, by learning the initial state of the SDM memory and therefore using it as a parametric memory, we show that the model further improves on a wide range of common-knowledge and reasoning tasks.
}

\metadata[Code]{\url{https://github.com/facebookresearch/sparse-delta-memory}}

\begin{document}

\maketitle

\section{Introduction}

\begin{minipage}{0.5\linewidth}
As frontier models 
continue to progress, they 
are leveraged in increasingly more complex tasks. In particular, the emergence of agentic settings involves sustained reasoning over long contexts.
This requires memory mechanisms that preserve long-range dependencies across extended interactions.
In standard transformer architectures equipped with vanilla softmax attention, the interaction memory is stored in a Key-Value (KV) cache.
Softmax attention excels at in-context learning, yet the KV cache, and therefore the compute and memory per token, all grow linearly with the sequence length (Fig.~\ref{fig:state_flops}, green line).
This unbounded growth is a limitation for in-context learning over very long sequences, such as entire codebases, extended reasoning traces, or videos, which play a central role in world modeling and robotics.
\medskip
\end{minipage}
%
\hfill
\begin{minipage}{0.42\linewidth}
\centering
\includegraphics[width=0.85\linewidth]{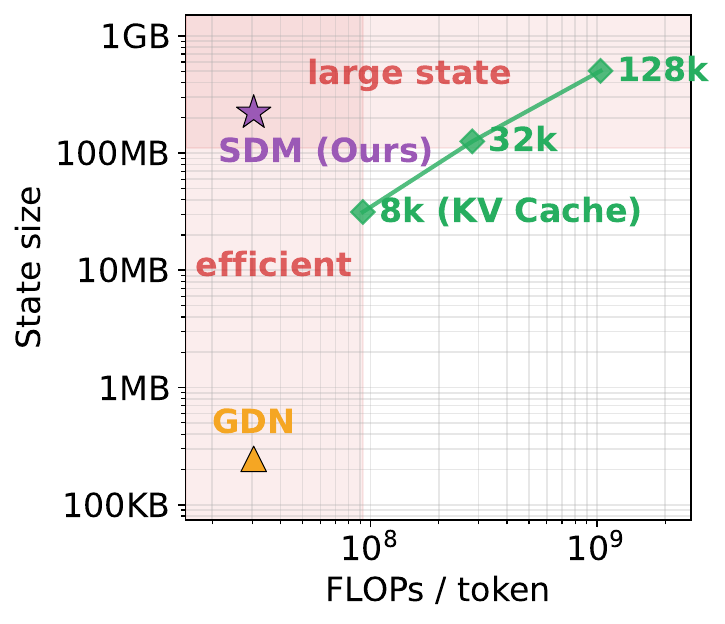}
\vspace{-1.0em}
\captionof{figure}{\textbf{State size vs FLOPs per token} (global layer, 1.4B). KV cache scales linearly with the sequence length. Our SDM approach offers a large state with constant FLOPs. 
\medskip
\label{fig:state_flops}}
\end{minipage}
%

A possible way to avoid this growing KV cache is to replace the explicit storage of all past tokens with a compressed recurrent state. Recurrent Neural Networks (RNNs), including State Space Models \citep{gu2024mambalineartimesequencemodeling} and Linear Attention variants \citep{katharopoulos2020transformersrnnsfastautoregressive, beck2024xlstmextendedlongshortterm}, rely on this strategy: they compress information into a fixed-size hidden state, maintaining constant memory and compute per token regardless of sequence length.
This enables the processing of arbitrarily long contexts without an explicit token limit.
However, their extremely small state sizes limit recall capability compared to transformers \citep{fu2023hungryhungryhipposlanguage}.
Indeed, \citet{arora2025simplelinearattentionlanguage} show that long-context performance is fundamentally bounded by the hidden state size.
Simply increasing the RNN memory size would improve recall, but modern linear RNNs like Mamba2~\citep{mamba2} and Gated DeltaNet (GDN)~\citep{yang2025gateddeltanetworksimproving} are bottlenecked by dense state updates that become prohibitively expensive as the state size grows.

To address the limitations of dense state updates, we introduce Sparse Delta Memory (SDM), a novel architecture based on the observation that the GDN update rule can be sparsified. This enables a three-order-of-magnitude increase in memory state size while maintaining the same compute budget, as shown in Fig.~\ref{fig:state_flops}.
Thanks to its larger state size, SDM significantly outperforms GDN on long-context recall tasks from the RULER~\citep{hsieh2024rulerwhatsrealcontext} benchmark and also shows better in-context learning capabilities on sequences of up to 1 million tokens.
Moreover, we show that contrary to GDN, learning the initial state of SDM allows the model to store meaningful pretraining knowledge that significantly improves performance on a wide range of metrics.
Indeed, in isoFLOP comparisons (Section~\ref{sec:scaling_laws}), an SDM with a learned initial state consistently achieves a lower training loss than GDN across all scales.
We validate our findings by training 8B activated parameter models on more than 1 trillion tokens.
Overall, we show that SDM, thanks to its large state, keeps the constant-space and memory advantages of Linear RNNs while significantly improving performance on long-context tasks which have been the main limitation of Linear RNNs like GDN and Mamba so far.
\ifneuripsversion An extended version with additional results is available on arXiv~\citep{cabannes2026sparsedeltamemoryscaling}.\fi

\section{Background}

\paragraph{Linear Attention as an Associative Memory.}
A fundamental perspective introduced by \citet{katharopoulos2020transformersrnnsfastautoregressive} is that attention writes outer products of keys $\mathbf{k}_t\in \mathbb{R}^{d_{\text{qk}}}$ and values $\mathbf{v_t} \in \mathbb{R}^{d_{\text{v}}}$ into a memory. Then, to read from the memory, a query $\mathbf{q_t}\in \mathbb{R}^{d_{\text{qk}}}$ is compared to the previous keys using a similarity metric, usually the inner product $\langle \mathbf{q}, \mathbf{k} \rangle$.
Moreover, to improve the accuracy of information retrieval, a pre-processing feature mapping $\phi$ can be applied to the keys and queries. We can then define the memory tensor $\mathbf{M}_t$ and normalization factor $\mathbf{z}_t$:
\[
\mathbf{M}_t \;=\; \sum_{i=1}^{t} \phi(\mathbf{k}_i)\, \mathbf{v}_i^\top \;\;\in\;\mathbb{R}^{d_{\text{qk}}\times d_{\text{v}}},\qquad
\mathbf{z}_t \;=\; \sum_{i=1}^{t} \phi(\mathbf{k}_i) \;\;\in\;\mathbb{R}^{d_{\text{qk}}}.
\]
A normalized read is then:
\begin{equation}
\mathbf{y}_t \;=\; \frac{\mathbf{M}_t^\top \phi(\mathbf{q}_t)}{\mathbf{z}_t^\top \phi(\mathbf{q}_t)}
\;=\; \frac{\sum_{i\le t} \langle \phi(\mathbf{q}_t),\phi(\mathbf{k}_i)\rangle\, \mathbf{v}_i}{\sum_{i\le t} \langle \phi(\mathbf{q}_t),\phi(\mathbf{k}_i)\rangle}.
\label{eq:la-read}
\end{equation}

Note that there exists an infinite-dimensional feature mapping $\phi$ such that  $ \langle \phi(\mathbf{q}_t),\phi(\mathbf{k}_i)\rangle\ = e^{ \langle \mathbf{q}_t,\mathbf{k}_i\rangle}$, in which case Eq. \eqref{eq:la-read} computes exactly softmax attention.
On the contrary, if $\phi$ is finite-dimensional, then the feature mapped keys and queries as well as the memory $\mathbf{M}_t$ are finite-dimensional and can thus be materialized and cached for future retrievals. That is, $(\mathbf{M}_t,\mathbf{z}_t)$ is stored in \emph{constant} memory, no matter the sequence length.

\paragraph{Gated DeltaNet (GDN).}
DeltaNet \citep{schlag2021lineartransformersselfretrievalcompressive} improves upon vanilla linear attention by introducing the \emph{delta rule}: before writing a new association into the memory, the model first retrieves and subtracts the existing value associated with the key, thereby preventing interference and keeping the memory norm bounded.
Gated DeltaNet (GDN) further adds a decay (forget) gate $\alpha_t \in (0,1)$ to control how quickly old associations are forgotten. This gives the following state update formula:
\begin{equation}
\mathbf{M}_t \leftarrow \alpha_t \, \mathbf{M}_{t-1} + \beta_t \, \mathbf{k}_t \left(\mathbf{v}_t - \alpha_t \mathbf{M}_{t-1}^\top \mathbf{k}_t\right)^\top,
\label{eq:gdn-update}
\end{equation}
where $\beta_t \in [0,1]$ is a learned input gate modulating the memory update strength.
In GDN, $\mathbf{k}_t$ and $\mathbf{q}_t$ are dense vectors in $\mathbb{R}^{d_{\text{qk}}}$, and the state $\mathbf{M}_t \in \mathbb{R}^{d_{\text{qk}} \times d_{\text{v}}}$ is a dense matrix.
The per-token computational cost is $O(d_{\text{qk}} \times d_{\text{v}})$.
Thus, increasing the size of the state represents a linear increase in FLOPs.

\paragraph{Product Key Memory.}
Product-Key Memories (PKM)~\citep{lample2019largememorylayersproduct} propose an indexing scheme that allows selecting $k$ arbitrary indices among $N$ possible memory slots, while scaling the number of computations sublinearly with respect to $N$.
Indeed, PKM can scale to memories of size $N \times d_{\text{v}}$ with $N \approx 10^6$ \citep{berges2024memorylayersscale}, while previous dense approaches are limited to $N \approx 10^3$ or $10^4$.
Given two sets of scores $\textbf{s}_1 \in \mathbb{R}^{\sqrt{N}}$ and $\textbf{s}_2 \in \mathbb{R}^{\sqrt{N}}$, one can get $N$ scores, one for each possible memory index, by doing the outer sum between the two score vectors:
$\textbf{s} \in \mathbb{R}^{\sqrt{N} \times \sqrt{N}} = \textbf{s}_1 \oplus \textbf{s}_2.$
We simply get the $\text{top}_k$ of the flattened scores $\textbf{s}$.
Since \mbox{$\text{top}_k (\textbf{s}_1 \oplus \textbf{s}_2) = \text{top}_k (\text{top}_k(\textbf{s}_1) \oplus \text{top}_k(\textbf{s}_2))$}, we do not  need to materialize the entire $N$ scores $\textbf{s}$ but only $k^2$ scores, making the indexing operation very efficient.
Since obtaining the scores $\mathbf{s_1}$ and $\mathbf{s_2}$ requires $O(\sqrt{N} \times d)$ operations and obtaining the final $\text{top}_k$ scores requires $O(k^2)$ operations, the indexing scheme of PKM has a $O(\sqrt{N} \times d + k^2)$ time complexity.
It is important to note that in PKM, the memory is not updated by the context and is learned only through training, analogous to the FFN weights in the transformer.

\section{Method}

\begin{figure}[t]
\centering
\input{assets/fig_sdm_overview}
\caption{\textbf{From GDN to SDM} (here $d_{\text{qk}}\,{=}\,8$, $N\,{=}\,16$). \textbf{(a)}~GDN writes $\mathbf{v}_t$, weighted by $\mathbf{k}_t$, to all $d_{\text{qk}}$ rows of its $d_{\text{qk}}\times d_{\text{v}}$ state. \textbf{(b)}~SDM scores each of the $N\,{=}\,(d_{\text{qk}}/2)^2$ slots with the outer sum of the key halves $\mathbf{k}'_{1,t}\oplus\mathbf{k}'_{2,t}$. \textbf{(c)}~The delta write reads and updates only the top-$W$ slots; reads select $R$ slots likewise. With $W\,{=}\,R\,{=}\,d_{\text{qk}}$, SDM touches as many slots per token as GDN does, matching FLOPs (Sect.~\ref{sec:isoflop}). See the detailed layer in Fig.~\ref{fig:sdm_architecture}.}
\label{fig:gdn_to_sdm}
\end{figure}

\subsection{Sparse Delta Memory}
\label{sec:sdm-method}

Our key insight is that the GDN update rule in Eq.~\eqref{eq:gdn-update} can be \emph{sparsified}: rather than applying the decay and delta update to the entire dense state $\mathbf{M}_t \in \mathbb{R}^{d_{\text{qk}} \times d_{\text{v}}}$, we maintain an explicit memory table $\mathbf{M}_t \in \mathbb{R}^{N \times d_{\text{v}}}$ with $N$ slots, and apply gated delta updates \emph{only to the $W$ slots selected by sparse keys}.
Concretely, at each timestep $t$ and for each SDM head, we propose:
\begin{enumerate}[leftmargin=*,itemsep=2pt]
\item \textbf{Sparse Key Selection.} Pre-PKM write keys $\mathbf{k'}_t$ and read queries $\mathbf{q'}_t$ are projected from the input $\mathbf{x}_t \in \mathbb{R}^d$ (model width $d$) via learned linear projections $W_k, W_q \in \mathbb{R}^{d \times 2\sqrt{N}}$. Each projected vector is split into two halves ($\mathbf{k}'_{1,t}, \mathbf{k}'_{2,t}$ and $\mathbf{q}'_{1,t}, \mathbf{q}'_{2,t} \in \mathbb{R}^{\sqrt{N}}$), and their outer sum $\mathbf{k}'_{1,t} \oplus \mathbf{k}'_{2,t} \in \mathbb{R}^{\sqrt{N} \times \sqrt{N}}$ yields $N$ scores: one per memory slot. Applying $\text{top}\text{-}W$ to the write scores and $\text{top}\text{-}R$ to the read scores selects the $W$ write indices $\mathcal{I}^w_t$ and $R$ read indices $\mathcal{I}^r_t$ from $N$ possible slots with $O(\sqrt{N} \times d + W^2 +R^2)$ compute. 

\item \textbf{Gated Delta Write.} Each selected write slot $i \in \mathcal{I}^w_t$ is first decayed by the forget gate $\alpha_t$:
\begin{equation}
\mathbf{\Tilde{M}}_t[i] \leftarrow \alpha_t \cdot \mathbf{M}_{t-1}[i]
\label{eq:sdm-decay}
\end{equation}
Following the delta rule, we then retrieve the value $\mathbf{\Tilde{v}}_t$ currently associated with the write key:
\begin{equation}
\mathbf{\Tilde{v}}_t = \mathbf{\Tilde{M}}_{t}^\top \mathbf{k}_t = \sum_{j \in \mathcal{I}^w_t} k_t^{(j)} \cdot \mathbf{\Tilde{M}}_{t}[j] \;\; \in \mathbb{R}^{d_{\text{v}}}
\label{eq:sdm-retrieve}
\end{equation}
and write the residual $\Delta \mathbf{v}_t = \mathbf{v}_t - \mathbf{\Tilde{v}}_t$ to each selected slot, weighted by that slot's key value:
\begin{equation}
\mathbf{M}_t[i] \leftarrow \mathbf{\Tilde{M}}_{t}[i] + \underbrace{\beta_t}_{\text{input gate}} \cdot k_t^{(i)} \cdot \underbrace{\left(\mathbf{v}_t - \mathbf{\Tilde{v}}_t\right)}_{\Delta \mathbf{v}_t}
\label{eq:sdm-write}
\end{equation}
where $\alpha_t = \exp( -A \cdot \text{softplus}(W_\alpha \mathbf{x}_t + b_{\text{dt}}))$ is a per-head forget gate and $\beta_t = \sigma(W_\beta \mathbf{x}_t)$ the input gate, $\mathbf{k}_t \in \mathbb{R}^N$ the sparse write-key vector (nonzero only on $\mathcal{I}^w_t$, entries $k_t^{(i)}$), $\mathbf{v}_t = W_v \mathbf{x}_t$ the value vector, and $A$, $b_{\text{dt}}$, $W_\alpha$, $W_\beta$ are learned ($\sigma$: sigmoid).
Unselected slots ($i \notin \mathcal{I}^w_t$) remain unchanged: $\mathbf{M}_t[i] = \mathbf{M}_{t-1}[i]$.

\item \textbf{Sparse Read.} The memory is then read by a weighted sum of the $R$ selected read slots:
\begin{equation}
\mathbf{y}_t = \mathbf{M}_t^\top \mathbf{q}_t = \sum_{i \in \mathcal{I}^r_t} q_t^{(i)} \cdot \mathbf{M}_t[i]
\label{eq:sdm-read}
\vspace{-0.5em}
\end{equation}
\item \textbf{Norm, Gating, and Head Mixing.} The retrieved memory $\mathbf{y}_t$ is normalized by a per-head LayerNorm, element-wise gated with $\mathbf{g}_t = \sigma(W_g \mathbf{x}_t) \in \mathbb{R}^d$, and finally a projection $W_o$ mixes the outputs from all SDM heads to produce the final layer output $\mathbf{o}_t \in \mathbb{R}^d$.
\end{enumerate}

\noindent\textbf{Connection to GDN.} When $N\,{=}\,d_{\text{qk}}$, $W\,{=}\,R\,{=}\,d_{\text{qk}}$ (all slots selected), and the sparse key values $k_t^{(i)}$ form a dense vector, $\mathbf{\Tilde{v}}_t\,{=}\,\alpha_t \mathbf{M}_{t-1}^\top \mathbf{k}_t$ and Eq.~\eqref{eq:sdm-write} recovers exactly the GDN update.
In this case, the only difference is the lack of 1D convolutions on the $q$$k$$v$ vectors, present in GDN but not in SDM.

\noindent\textbf{Learned Initial State $\mathbf{M_0}$.} Compared to GDN, which has a very small state size, SDM has a much larger state. This property might not serve only as a storage mechanism for in-context knowledge, but also, if one considers $\mathbf{M_0}$ as a learnable parameter of the model, the SDM memory can learn knowledge during pretraining and reuse it at test time. Since having a learned $\mathbf{M_0}$ does not add any FLOPs at inference time compared to a null-initialized $\mathbf{M_0}$, we chose the learned $\mathbf{M_0}$ variant as the default setting for SDM.
We ablate the impact of learning $\mathbf{M_0}$ in Section~\ref{sec:ablations}.

\noindent\textbf{Efficient Training.} We detail how SDM is trained efficiently in Appendix~\ref{sec:efficient-training}.

\subsection{IsoFLOP Design: Matching GDN Parameters and FLOPs}
\label{sec:isoflop}
%
We ensure that SDM has the same number of parameters and FLOPs as the dense GDN baseline by using an isoFLOP setting in all comparisons. Hence, any improvement stems from the larger memory capacity alone.

\noindent\textbf{Parameters.} Both GDN and SDM share identically-sized linear projections: $W_q, W_k \in \mathbb{R}^{d \times d_{\text{qk}}^{\text{total}}}$ and $W_v \in \mathbb{R}^{d \times d_{\text{v}}^{\text{total}}}$, where $d_{\text{qk}}^{\text{total}}=H^{\text{GDN}} \times d_{\text{qk}}^{\text{GDN}}=d/2$ and $d_{\text{v}}^{\text{total}}=H^{\text{GDN}} \times d_{\text{v}}^{\text{GDN}}=d$.

\noindent\textbf{FLOPs.} GDN's per-token cost is $O(H^{\text{GDN}} \times d_{\text{qk}}^{\text{GDN}} \times d_{\text{v}}^{\text{GDN}})=O(d_{\text{qk}}^{\text{GDN}} \times d_{\text{v}}^{\text{total}})$ because every head accesses its full $d_{\text{qk}} \times d_{\text{v}}$ state. SDM's cost is instead $O(H^{\text{SDM}} \times (W{+}R) \times d_{\text{v}}^{\text{SDM}})=O((W{+}R) \times d_{\text{v}}^{\text{total}})$, independent of the memory size $N$. Setting $W=R=d_{\text{qk}}^{\text{GDN}}$ thus yields matching FLOPs. To be precise, PKM's top-k on the outersum of scores is an additional computation, but it amounts to less than 1\% of the layer's FLOPs.

We set $d_{\text{qk}}^{\text{GDN}}\,{=}\,64$, $d_{\text{v}}^{\text{GDN}}\,{=}\,128$, giving $W_q, W_k \in \mathbb{R}^{d \times d/2}$ and $W_v \in \mathbb{R}^{d \times d}$ for both GDN and SDM.

\subsection{Limiting the State Size Expansion}
\label{sec:state-size}
With SDM (and unlike GDN), fewer heads do not increase FLOPs but do increase the memory size, which improves performance \citep{arora2025simplelinearattentionlanguage}. Having a single head under a parameter constraint maximizes the size of the memory.
Indeed, the memory size of SDM scales as follows:
\begin{equation}
M_{\text{size}} = H \times \left[\frac{d_{\text{qk}}^{\text{total}}}{2H} \times \frac{d_{\text{qk}}^{\text{total}}}{2H} \times \frac{d_{\text{v}}^{\text{total}}}{H}\right] = \frac{(d_{\text{qk}}^{\text{total}})^2 \cdot d_{\text{v}}^{\text{total}}}{4H^2}.
\end{equation}
However,
the total memory size per layer of SDM would scale as $O(d^3)$, faster than GDN's state growth ($O(d)$ with GDN heads and $O(d^2)$ without heads) and even faster than the model's parameter count ($O(d^2)$).
This would make the sparse memory impractically large at scale.
Therefore, H serves as a hyper-parameter with no impact on FLOPs but allowing one to control the state size in SDM.
\ifmultifactor

\noindent\textbf{Beyond Two Factors.} The two-factor product key used so far is the direct application of PKM~\citep{lample2019largememorylayersproduct}, but SDM's addressing extends to $F$ factors: splitting each head's key into $F$ sub-keys of sizes $n_1,\dots,n_F$ addresses $N=\prod_f n_f$ slots with only $\sum_f n_f$ key dimensions. The number of factors thus decouples the state size from the parameters spent on addressing, which makes many heads affordable at a constant state size. In a 1.4B comparison, 16 heads with four factors have lower perplexity than the two-head, two-factor configuration of this paper at all evaluated context lengths up to 1M tokens, with the same state size and sparse read and write cost and 3\% wider key projections (Appendix~\ref{sec:app_multifactor}). The main results of this paper nonetheless use the simpler two-factor addressing.
\fi

\section{Experimental Setup}

\begin{table}[t]
\caption{\textbf{Scaling ladder configurations.} Params include embedding weights but exclude the SDM memory state. The State columns correspond to the total state size across all global layers, with St:Param reporting this size as a fraction of the non-embedding parameters.
\label{tab:ladder}}
\smallskip
\centering
\small
\SDMfitwidetable{\begin{tabular}{c|ccccc|cc|ccc}
\toprule
 & & & & & & \multicolumn{2}{c|}{GDN} & \multicolumn{3}{c}{SDM} \\
Level & $d$ & Layers & Params & Tokens & LR (${\times}10^{-3}$) & State & St:Param & $H$ & State & St:Param \\
\midrule
1 & \sz768  & 9  & 257M  & \sz10.7B  & 1.78 & \sz98k  & 0.16\% & 1 & \sz57M  & \sz94\% \\
2 & \sz768  & 11 & 280M  & \sz14.9B  & 1.50 & \sz98k  & 0.12\% & 1 & \sz57M  & \sz68\% \\
3 &    1024 & 11 & 407M  & \sz24.3B  & 1.35 & 131k    & 0.09\% & 1 & 134M    & \sz93\% \\
4 &    1024 & 14 & 465M  & \sz34.0B  & 1.24 & 197k    & 0.10\% & 1 & 201M    & \sz99\% \\
5 &    1280 & 14 & 658M  & \sz52.8B  & 1.20 & 246k    & 0.07\% & 1 & 393M    & 119\% \\
6 &    1536 & 15 & 847M  & \sz81.0B  & 0.99 & 295k    & 0.07\% & 1 & 679M    & 150\% \\
8 &    1920 & 21 & 1.48B & 168.4B    & 0.87 & 614k    & 0.06\% & 2 & 553M    & \sz56\% \\
\midrule
13 & 3840 & 38 & 8.14B & 1{.}141T & 0.50 & 2.2M & 0.03\% & 2 & 7.963B\sz  & 111\% \\
\bottomrule
\end{tabular}}
\end{table}

\paragraph{Architecture.}
All models use a hybrid architecture with Multi-Head Attention (MHA) layers using Sliding Window Attention (SWA) and interleaved global receptive field layers in a 3:1 short:long ratio.
To avoid harmful competition between the short and long layers \citep{cabannes2025shortwindowattentionenables}, we choose a small window size of 128 tokens which has been adopted for similar reasons in \citet{coreteam2026mimov2flashtechnicalreport} and in \citet{openai2025gptoss120bgptoss20bmodel}.
All full-attention layers use grouped-query attention (GQA; $n_h$ query, $n_{kv} = n_h/2$ key/value heads) and gated attention output \citep{qiu2025gatedattentionlargelanguage}.
Each block uses a gated MLP \citep{liu2021payattentionmlps} with SiLU activation \citep{elfwing2017sigmoidweightedlinearunitsneural} and a hidden dimension made to match the parameters of an equivalent non-gated MLP with a hidden dimension of $4d$.
Position encoding uses RoPE \citep{su2023roformerenhancedtransformerrotary} with base $\theta = 500{,}000$.

\noindent\textbf{SDM Configuration.}
The SDM layers use $W = R = 64$ reads and writes, and $N = (d/4H)^2$ memory slots (Section~\ref{sec:isoflop}).
Read and write activations use softmax-normalization.
The forget gate parameter $A$ is initialized uniformly in $[0, 16]$ and the time-step bias $b_{\text{dt}}$ is initialized from $\text{softplus}^{-1}(u)$, $u \sim \mathcal{U}[0.001, 0.1]$, both matching GDN/FLA conventions.
The number of SDM heads $H$ is set per scale to keep the state-to-parameter ratio around 1:1 (see Section~\ref{sec:state-size} and Table \ref{tab:ladder}).

\noindent\textbf{GDN and Mamba2 Baselines.}
The GDN and Mamba2 baselines use $d_{\text{qk}} = 64$, $d_{\text{v}} = 128$, and $n_h$ heads, matching the number of attention heads.

\noindent\textbf{Training. }
All models were pretrained on 8192-token sequences of diverse text data.
Training follows a Warmup–Stable–Decay (WSD) LR schedule with gradient clipping at 1.0 using the AdamW optimizer \citep{loshchilov2019decoupledweightdecayregularization} with $\beta_1\,{=}\,0.9$ and $\beta_2\,{=}\,0.95$.
Learning rates were tuned for the transformer baseline and confirmed optimal for GDN via small grid searches:
we use a uniform LR across all architectures, varying only by model scale (Table~\ref{tab:ladder}).
To maximize recall performance, 1.4B and 8B models undergo a long-context fine-tuning stage on 128k-token sequences using 4B and 16B tokens, respectively.

\noindent\textbf{Scaling Ladder.}
To evaluate the capability of our approach to scale at larger scale, we train a ladder of model sizes, where we adopt a 160 tokens-per-parameter (160TPP) compute budget. The architectures considered for the ladder are FullAttn, GDN and SDM.
We do not rely on an optimal token budget following the Chinchilla scaling laws \citep{hoffmann2022trainingcomputeoptimallargelanguage} because most architectures and models currently deployed at scale are trained way beyond their training-optimal compute budget.
We thus choose this ${\sim}160$ tokens per non-embedding parameter budget to compare the architectures in a rather inference-optimal setting.
Except for the head number $H$, GDN, SDM and FullAttn share identical hyperparameters at each level.

\noindent\textbf{Evaluation.} We evaluate models using validation NLL on held-out natural text and coding data. We also evaluate them on a diverse set of reasoning and commonsense tasks as listed in Table~\ref{tab:l08_accuracy}.

\section{Results}

\begin{figure}[t]
    \centering
    \includegraphics[width=0.7\linewidth]{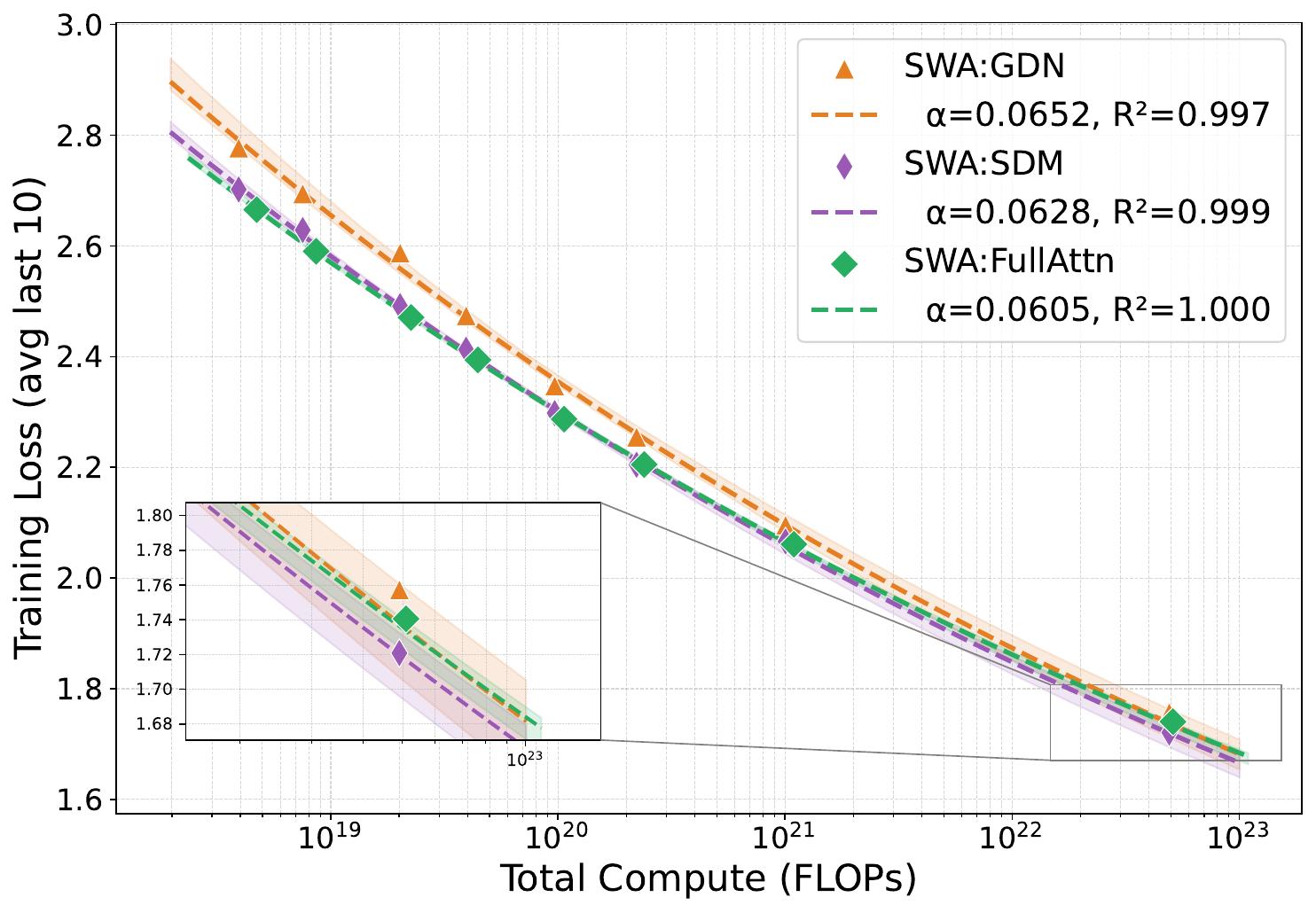}
    \caption{\textbf{Scaling laws: loss vs FLOPs.} Average of last 10 log entries vs total compute.  SDM outperforms GDN across scales. As predicted by the scaling law, it slightly outperforms full attention at 8B scale.
    \label{fig:train_loss_scaling}}
\end{figure}

In the following sections, we report SDM performance compared to GDN, Mamba2 and FullAttn as global layers.
First, we examine scaling laws across the scaling ladder after pre-training (Section~\ref{sec:scaling_laws}).
Second, we report the performance on short- and long-context tasks of the long-context finetuned models (Sections \ref{sec:short-context}, \ref{sec:ruler}, \ref{sec:code_ppl}).
Last, we perform extensive ablations verifying SDM's utility and memory use (Section~\ref{sec:ablations}).

\subsection{Power-Law Scaling: SDM Outperforms GDN at All Compute Levels}
\label{sec:scaling_laws}

We start by evaluating SDM's compute efficiency by measuring how training loss decreases as total training compute (FLOPs) increases.
Specifically, we compare SDM against GDN at each level of the scaling ladder (Table~\ref{tab:ladder}), keeping both architectures matched in FLOPs and parameters (excluding SDM's sparse embedding memory).
We used the final losses of the levels from 1 to 8 to compute scaling laws and predict the loss at 8B (level 13) scale.
As shown in Figure \ref{fig:train_loss_scaling}, SDM outperforms GDN at every level of the scaling ladder. Moreover, SDM provides predictable scaling with a coefficient of determination $R^2=0.999$.
As predicted by our scaling law, when trained at level-13 (8B parameters) scale, SDM reaches a significantly lower loss than GDN (which slightly underperforms its predicted loss but stays within the 95\% confidence interval computed using bootstrapping) and even outperforms the 8B model with FullAttn.
These results demonstrate that SDM consistently outperforms an iso-FLOP GDN across all compute levels while remaining competitive or even slightly surpassing FullAttn at larger scales.

\begin{table}[t]
\centering
{\small
\caption{\textbf{Performance of different global layer choices} at 1.4B (L08) and 8B (L13) scale. All models use SWA as the local layer. Delta column shows delta in performance between SDM and GDN. Bold marks the best model at each scale. $^\dagger$BoolQ is the least stable task in the suite (Appendix~\ref{sec:app_boolq}).
\label{tab:l08_accuracy}}
\smallskip
\SDMfitwidetable{\begin{tabular}{@{\ }lr|rrr|r|rr@{\ }}
\toprule
& \multicolumn{4}{c|}{1.4B (L08)} & \multicolumn{3}{c}{8B (L13)} \\
\multicolumn{1}{r}{Local layer $\rightarrow$} & \multicolumn{4}{c|}{SWA} & \multicolumn{3}{c}{SWA} \\
\multicolumn{1}{r}{Global layer $\rightarrow$} & FullAttn & Mamba2 & GDN & SDM\quad~ & FullAttn & GDN & SDM\quad~ \\
\midrule
Validation text NLL {\small$\downarrow$} & 2.660 & 2.686 & 2.683 & \textbf{2.654} \subg{-0.029} & 2.285 & 2.298 & \textbf{2.253} \subg{-0.045} \\
\midrule
HellaSWAG~\evalcite{zellers2019hellaswagmachinereallyfinish} {\small$\uparrow$} & 61.47 & 61.31 & 61.27 & \textbf{62.04} \subg{+0.77} & 79.33 & 79.10 & \textbf{80.02} \subg{+0.92} \\
WinoGrande~\evalcite{sakaguchi2019winograndeadversarialwinogradschema} {\small$\uparrow$} & 60.54 & \textbf{62.67} & 61.56 & 62.12 \subg{+0.56} & 73.64 & 73.24 & \textbf{75.30} \subg{+2.06} \\
ARC easy~\evalcite{clark2018thinksolvedquestionanswering} {\small$\uparrow$} & 61.56 & 62.88 & 62.49 & \textbf{63.00} \subg{+0.51} & 78.10 & \textbf{79.15} & 78.52 \subr{-0.63} \\
ARC challenge~\evalcite{clark2018thinksolvedquestionanswering} {\small$\uparrow$} & 33.56 & 34.08 & 34.59 & \textbf{34.76} \subg{+0.17} & 50.82 & 52.27 & \textbf{53.05} \subg{+0.78} \\
PIQA~\evalcite{bisk2019piqareasoningphysicalcommonsense} {\small$\uparrow$} & 73.83 & 73.67 & 73.83 & \textbf{74.27} \subg{+0.44} & 79.98 & \textbf{81.01} & 80.47 \subr{-0.54} \\
OpenBookQA~\evalcite{OpenBookQA2018} {\small$\uparrow$} & 35.80 & 36.40 & \textbf{36.80} & 36.00 \subr{-0.80} & 43.80 & 43.00 & \textbf{44.60} \subg{+1.60} \\
RACE.mid~\evalcite{lai2017racelargescalereadingcomprehension} {\small$\uparrow$} & \textbf{53.55} & 50.28 & 50.42 & 49.65 \subr{-0.77} & \textbf{64.83} & 60.17 & 62.05 \subg{+1.88} \\
RACE.high~\evalcite{lai2017racelargescalereadingcomprehension} {\small$\uparrow$} & \textbf{39.62} & 36.25 & 36.68 & 37.76 \subg{+1.08} & \textbf{47.94} & 43.40 & 44.97 \subg{+1.57} \\
CommonsenseQA~\evalcite{talmor2019commonsenseqaquestionansweringchallenge} {\small$\uparrow$} & 19.57 & 20.31 & 19.33 & \textbf{20.72} \subg{+1.39} & 68.88 & 66.83 & \textbf{70.60} \subg{+3.77} \\
BoolQ$^\dagger$~\evalcite{clark2019boolqexploringsurprisingdifficulty} {\small$\uparrow$} & 62.42 & 62.75 & \textbf{63.03} & 62.78 \subr{-0.25} & 71.83 & \textbf{74.98} & 67.13 \subr{-7.85} \\
TQA~\evalcite{joshi-etal-2017-triviaqa} {\small$\uparrow$} & 24.24 & 23.48 & 23.68 & \textbf{26.61} \subg{+2.93} & 55.23 & 55.36 & \textbf{59.11} \subg{+3.75} \\
HumanEval+/pass@1~\evalcite{evalplus} {\small$\uparrow$} & 8.54 & \textbf{9.76} & 8.54 & \textbf{9.76} \subg{+1.22} & \textbf{24.39} & 18.29 & \textbf{24.39} \subg{+6.10} \\
NaturalQuestions~\evalcite{kwiatkowski-etal-2019-natural} {\small$\uparrow$} & 7.70 & 8.09 & 7.76 & \textbf{8.23} \subg{+0.47} & 22.94 & 22.60 & \textbf{25.93} \subg{+3.33} \\
MMLU~\evalcite{hendryckstest2021} {\small$\uparrow$} & 24.54 & 25.03 & 26.02 & \textbf{27.24} \subg{+1.22} & \textbf{58.73} & 57.24 & 57.81 \subg{+0.57} \\
GSM8K~\evalcite{cobbe2021gsm8k} {\small$\uparrow$} & 2.81 & \textbf{3.03} & 2.96 & 2.73 \subr{-0.23} & \textbf{29.34} & 28.81 & 28.66 \subr{-0.15} \\
\midrule
Average accuracy {\small$\uparrow$} & 37.98 & 38.00 & 37.93 & \textbf{38.51} \subg{+0.58} & 56.65 & 55.70 & \textbf{56.84} \subg{+1.14} \\
\midrule
\multicolumn{8}{@{\ }l}{\textit{RULER long-context tasks (exact match, averaged over lengths 4k--131k)}} \\
single 1 {\small$\uparrow$} & 64.2 & 63.5 & 99.9 & \textbf{100.0} \subg{\sz+0.1} & 99.3 & \textbf{100.0} & \textbf{100.0}\subg{\sz+0.0} \\
single 2 {\small$\uparrow$} & 53.1 & 28.2 & 20.7 & \textbf{70.8} \subg{+50.1} & \textbf{89.4} & 45.1 & 71.5\subg{+26.4} \\
single 3 {\small$\uparrow$} & 41.1 & 13.4 & 12.1 & \textbf{46.3} \subg{+34.2} & 70.1 & 32.6 & \textbf{74.9}\subg{+42.3} \\
multikey 1 {\small$\uparrow$} & \textbf{44.0} & 14.7 & 13.6 & 35.0 \subg{+21.4} & \textbf{75.3} & 32.9 & 59.3\subg{+26.4} \\
multikey 2 {\small$\uparrow$} & \textbf{25.3} & 0.7 & 0.7 & \sz0.8 \subg{\sz+0.1} & \textbf{58.0} & 1.0 & \sz4.7\subg{\sz+3.7} \\
multikey 3 {\small$\uparrow$} & \textbf{8.6} & 0.2 & 0.1 & \sz0.2 \subg{\sz+0.1} & \textbf{29.8} & 0.1 & \sz1.2\subg{\sz+1.1} \\
multivalue {\small$\uparrow$} & \textbf{39.8} & 17.5 & 11.7 & 37.4 \subg{+25.7} & \textbf{74.3} & 31.1 & 66.1\subg{+35.0} \\
multiquery {\small$\uparrow$} & 39.0 & 15.0 & 11.8 & \textbf{41.1} \subg{+29.3} & \textbf{73.0} & 31.2 & 68.6\subg{+37.4} \\
vt {\small$\uparrow$} & \textbf{32.1} & 8.3 & 16.6 & 23.7 \subg{\sz+7.1} & 67.2 & 46.2 & \textbf{72.3}\subg{+26.1} \\
cwe {\small$\uparrow$} & 5.8 & 3.6 & 6.6 & \textbf{8.7} \subg{\sz+2.1} & 5.5 & 11.4 & \textbf{12.8}\subg{\sz+1.4} \\
fwe {\small$\uparrow$} & 30.9 & \textbf{37.5} & 35.6 & 12.3 \subr{-23.3} & \textbf{77.8} & 62.7 & 65.0\subg{\sz+2.3} \\
qa 1 {\small$\uparrow$} & \textbf{19.3} & 11.5 & 13.3 & 13.3 \subg{\sz+0.0} & \textbf{39.0} & 23.6 & 27.2\subg{\sz+3.6} \\
qa 2 {\small$\uparrow$} & \textbf{20.9} & 18.5 & 18.3 & 18.0 \subr{\sz-0.3} & \textbf{38.4} & 28.2 & 30.4\subg{\sz+2.2} \\
\midrule
RULER average {\small$\uparrow$} & \textbf{32.6} & 17.9 & 20.1 & 31.4 \subg{+11.3} & \textbf{61.3} & 34.3 & 50.3 \subg{+16.0} \\
\bottomrule
\end{tabular}}}
\end{table}

\subsection{Sparse Memory Improves Short-Context Performance}
\label{sec:short-context}
We evaluate how SDM's large sparse memory impacts performance on reasoning and knowledge tasks that do not specifically involve large context capabilities, referred to as ``short context''. At both 1.4B and 8B scales, SDM achieves lower NLL and higher average accuracy than GDN; see Table~\ref{tab:l08_accuracy}. Notably, SDM obtains the lowest NLL among all models at both scales, outperforming even FullAttn. At both 1.4B and 8B scales, SDM improves over GDN on 11/15 tasks and achieves an average accuracy higher than GDN and comparable to FullAttn. 
This underscores the advantage of having a large state containing not only in-context information but also pretraining knowledge in the form of a learned initial state compared to the states of GDN or FullAttn which only store in-context knowledge and not pre-training knowledge.

\subsection{Scaling Memory Improves Long-Context Retrieval}

\label{sec:ruler}
Averaged over all lengths 4k--131k (see App.~\ref{sec:ruler-per-task} for per-length results), SDM is by far the strongest fixed-state model at both 1.4B and 8B scales; it improves on GDN on 10 of the 13 tasks at 1.4B and matches or improves on all 13 at 8B (Table~\ref{tab:l08_accuracy}).
FullAttn stays ahead overall (32.6 and 61.3) thanks to its unbounded KV cache, though SDM matches or exceeds it on 5 of the 13 tasks at 1.4B and 4 at 8B.
Fixed-state models struggle most on the dense multi-key tasks where full attention dominates.
A controlled probe (App.~\ref{sec:app_keytype}) separates two causes: at fixed state size, making the keys maximally distinct lifts SDM's exact-match accuracy roughly tenfold, so part of the gap is collision in the product-key address space and is mitigable by addressing and training; but at fixed keys, accuracy still orders by state size, so the remainder is the cost of compressing an unbounded context into a bounded state. Dense multi-key retrieval is therefore an open problem for fixed-state models in general.
Overall, RULER results demonstrate that the larger state size of SDM significantly improves long-context recall compared to previous linear RNNs. 

\subsection{SDM Reaches Lower Perplexity with More Context}
\label{sec:code_ppl}

\begin{figure}[t]
~ \hfill
\begin{minipage}{0.45\linewidth}
    \includegraphics[width=0.92\linewidth]{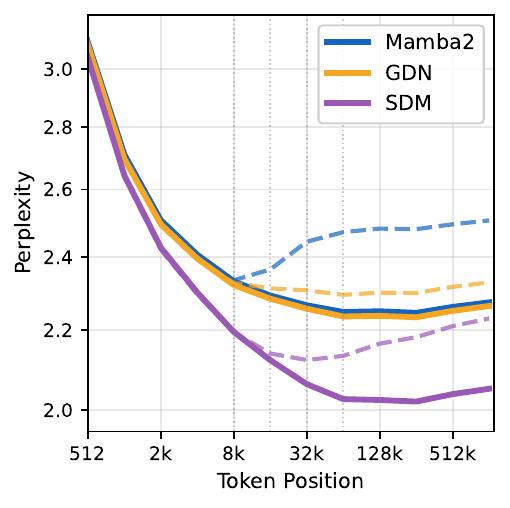}
    \captionof{figure}{\textbf{Perplexity by token position on code data} (1M token documents, 1.4B model size). Solid: 128k long-context finetuned; dashed: pre-trained. Local layers are SWA for all models.
    }
    \label{fig:ppl_by_position}
\end{minipage}
\hfill \hfill \hfill
\begin{minipage}{0.41\linewidth}
    \includegraphics[width=\linewidth]{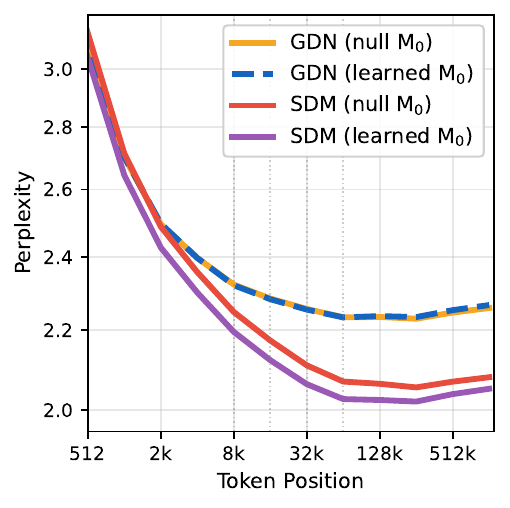}
    \caption{\textbf{Learned initial state ablation on code data }(1M token documents, 1.4B model size, post-trained). PPL by token position. Learning $\mathbf{M_0}$ benefits SDM but not GDN.
    \label{fig:ppl_by_position_ablation}}
\end{minipage}
\hfill ~
\end{figure}

SDM achieves consistently lower perplexity on code data compared to Mamba2 and GDN (Fig.~\ref{fig:ppl_by_position}).
Even at short sequences (512 tokens), SDM outperforms both baselines, a benefit we attribute to the learned initial state $\mathbf{M_0}$.
The advantage grows substantially at long contexts (32k–512k tokens), where SDM's perplexity decreases to near 2.0 while Mamba2 and GDN remain around 2.2–2.3.
On this evaluation data, the validation perplexity of all models seems to increase for token positions beyond 256k.
However, this is an artifact of the local, token-level perplexity being higher at those positions.
When we measure only the perplexity gain contributed by the long-context layers, SDM continues to improve beyond 256k tokens, up to 1 million tokens.
This highlights SDM's key strength: its large sparse memory retains information across extremely long sequences, whereas fixed-size state methods suffer from capacity constraints.

\section{Ablations: What Makes SDM Work?}
\label{sec:ablations}
\vspace{-0.6em}
\paragraph{Disentangling Memory Capacity from Learned Initialization.}
To verify that SDM's gains stem from increased memory capacity rather than the learned initial state $\mathbf{M_0}$, we ablate both components.
Figure \ref{fig:ppl_by_position_ablation} shows that SDM without a learned $\mathbf{M_0}$ substantially outperforms GDN, confirming that the state size is the primary driver of performance.
Adding a learned $\mathbf{M_0}$ to a vanilla GDN does not measurably improve performance. This is not surprising considering the limited state size.
For detailed results, see Table~\ref{tab:sdm_ablations}.

\vspace{-0.6em}
\paragraph{Impact of Memory State Size.}
\label{sec:mem-size-ablation}

\begin{table}[t]
\vspace{-0.5em}
\begin{minipage}[t]{0.66\linewidth}
    \footnotesize
    \setlength{\tabcolsep}{3pt}
    \renewcommand{\arraystretch}{0.95}
    \SDMfitwidetable{\begin{tabular}{@{}lll@{\ \ }cccc@{\ }}
        \toprule
         & Model & \textbf{$\mathbf{M_0}$} & State & Code NLL $\downarrow$ & Acc$\uparrow$  & RULER$\uparrow$  \\
        \midrule
\multirow{4}{*}{
\rotatebox{20}{
\begin{minipage}{2cm}
\it
ablation: learned \\
initial state (1.4B)
\end{minipage}}}
                   & GDN & null    & 0.6M      & 0.849         & 37.9         & 20.1 \\
                   & GDN & learned & 0.6M      & 0.850         & 38.0         & 20.6 \\
                   & SDM & null    & 553M      & 0.845         & 37.3         & 28.0 \\
                   & SDM & learned & 553M      & \textbf{0.822} & \textbf{38.5} & \textbf{31.4} \\
                   \midrule
\multirow{4}{*}{
\rotatebox{20}{
\begin{minipage}{2cm}
\it ablation: memory \newline size (0.8B)
\end{minipage}}}

                   & GDN & null    & 0.3M      & 0.963         & 33.8         & 16.0 \\
                   & SDM & learned & 42M       & 0.947         & 33.7         & 20.2 \\
                   & SDM & learned & 170M      & 0.937         & 33.6         & 20.7 \\
                   & SDM & learned & 679M      & \textbf{0.914} & \textbf{34.6} & \textbf{21.5} \\
        \bottomrule
    \end{tabular}}
\end{minipage}
\hfill
\begin{minipage}{0.31\linewidth}
\medskip
    \caption{\textbf{SDM ablations.} \textit{Top}: learned initial state ablation (L08, 1.4B). Making the memory $\mathbf{M}_0$ a learnable parameter improves code-data NLL, accuracy (\%), and RULER recall (\%). \textit{Bottom}: memory size ablation (L06). SDM degrades gracefully when memory is reduced; even at 42M it outperforms GDN on RULER.}
    \label{tab:sdm_ablations}
\end{minipage}
\vspace{-0.5em}
\end{table}

To confirm that memory size drives SDM's long-context advantage, we ablate the number of memory slots $N$ by varying the PKM key dimension $d_{\text{qk}}$ (which controls $N\,{=}\,(d_{\text{qk}}/2)^2$) while keeping $W$, $R$, $d_{\text{v}}$, and all non-SDM layers fixed.
Table~\ref{tab:sdm_ablations} (bottom) shows a monotonic NLL degradation as memory shrinks, confirming that a larger memory state improves modeling.
All SDM variants outperform GDN on long-context recall and show monotonic improvement with larger state sizes.

\paragraph{Efficiency.}

\begin{figure}[t]
\centering
\includegraphics[width=\linewidth]{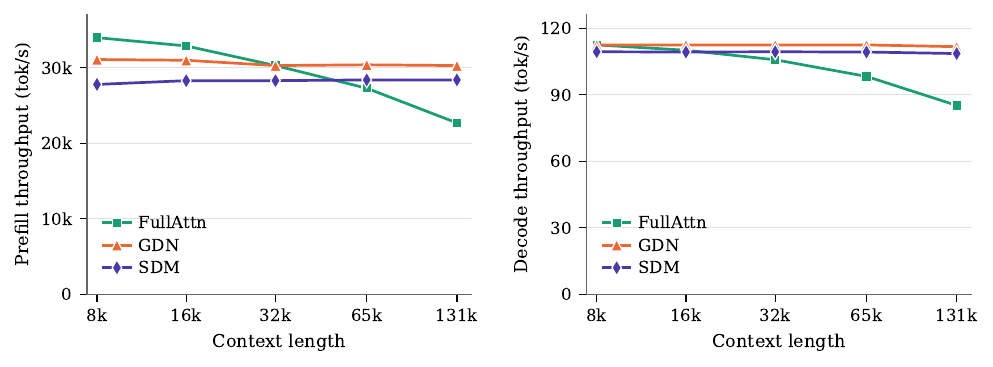}

\vspace{-0.2em}
\caption{\textbf{Inference throughput at 8B (L13)}, single H100, bf16, batch size 1. FullAttn
degrades with context in both regimes; the fixed states of GDN and SDM make them
flat in length.}
\label{fig:efficiency}
\end{figure}

Sparse approaches match dense ones in FLOPs but not  in memory traffic: GDN's small
state fits in SRAM, whereas SDM's lives in HBM. At 8B scale, a training step takes 2.516\,s for SDM against 2.177\,s for GDN and 2.004\,s
for FullAttn, a 16\% overhead over GDN for a state ${\sim}4{,}000\times$ larger
(Appendix~\ref{sec:app_compute_resources}).
For inference throughput, Figure~\ref{fig:efficiency} shows that SDM preserves
the near-flat profile of RNNs, while FullAttn degrades with longer sequences. 
SDM thus overtakes FullAttn beyond
$\sim$32k tokens in decode and $\sim$54k in prefill, reaching $1.3\times$ and $1.2\times$ its
throughput at 131k, where GDN leads SDM by 7\% in prefill and by only 3\% in
decode. 
SDM's sparsity is thus not merely theoretical; it is heavily leveraged by our implementation.

\begin{figure}[t]
    \centering
    \begin{subfigure}[h]{0.48\textwidth}
        \centering
        \includegraphics[width=\linewidth]{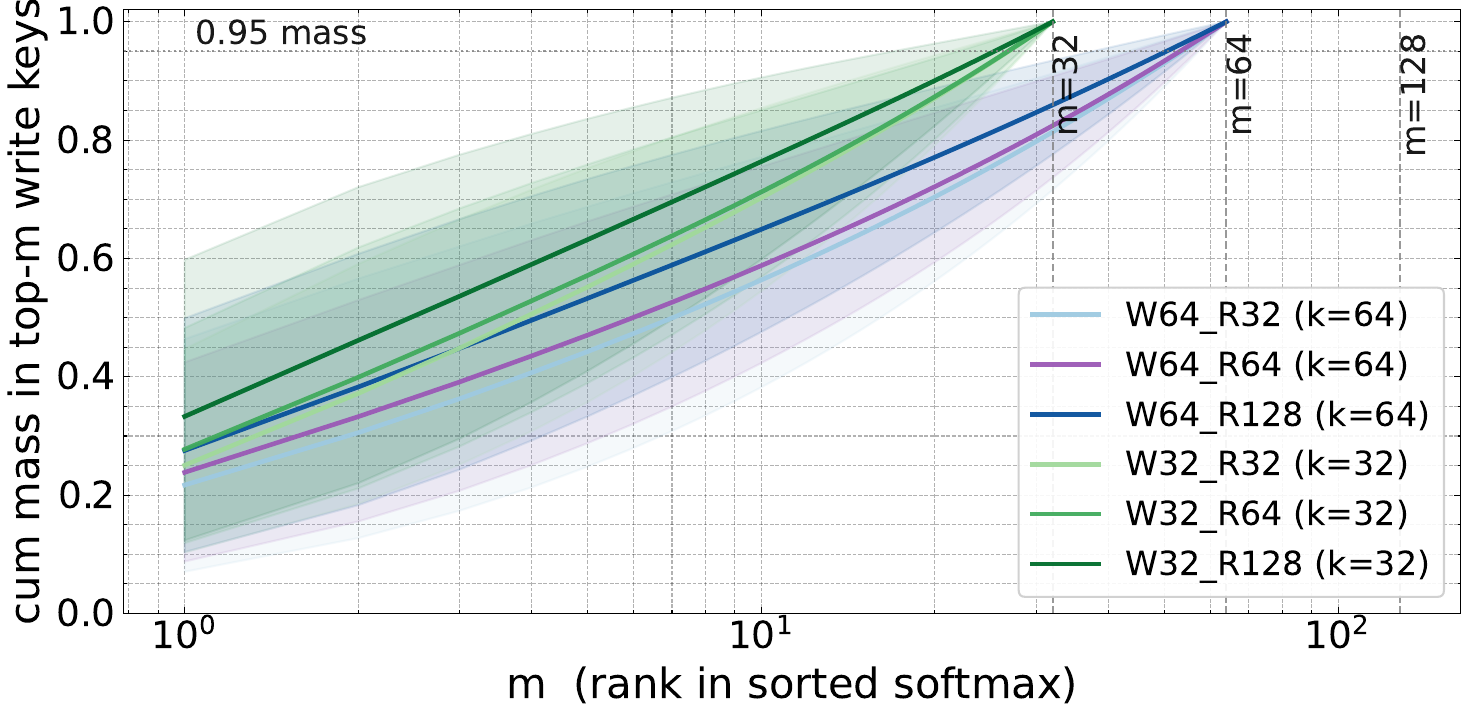}
        \caption{Write keys}
        \label{fig:topk-write}
    \end{subfigure}
    \hfill
    \begin{subfigure}[h]{0.48\textwidth}
        \centering
        \includegraphics[width=\linewidth]{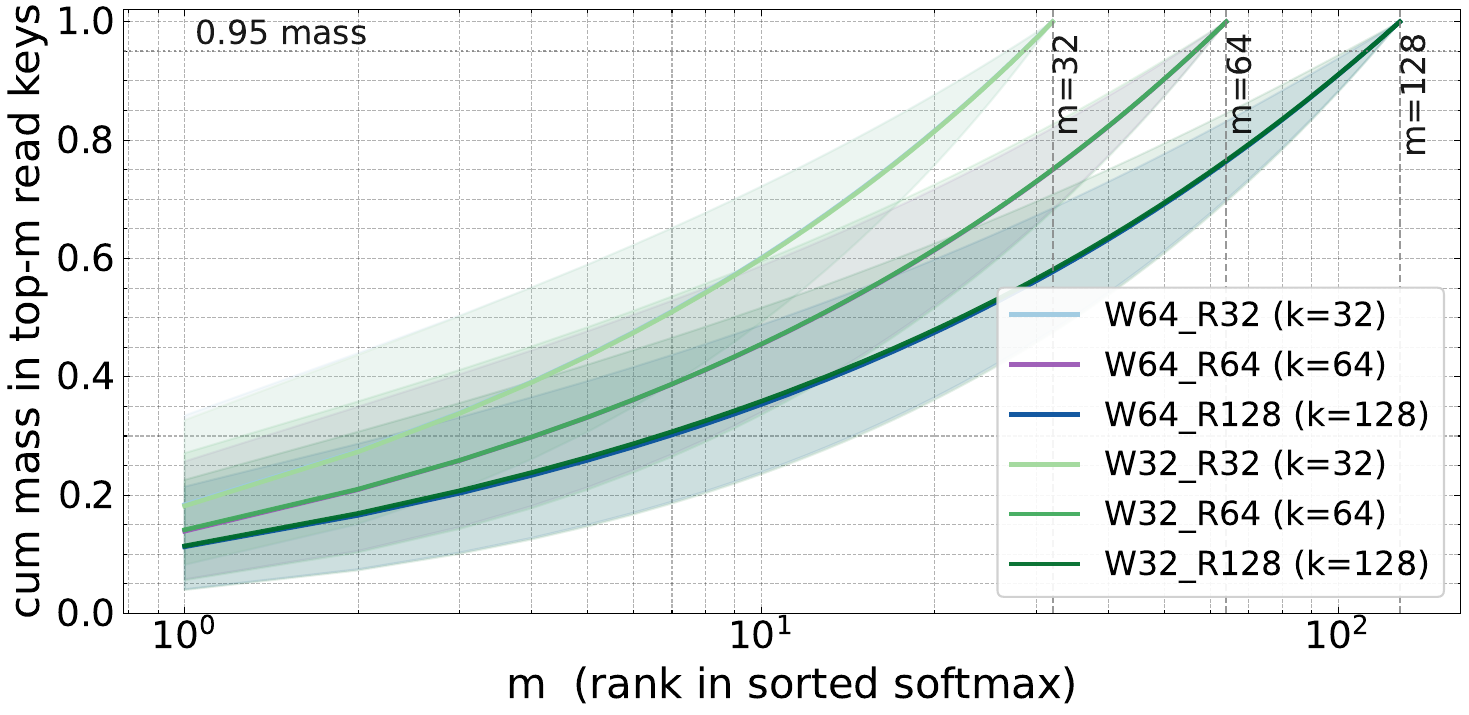}
        \caption{Read keys}
        \label{fig:topk-read}
    \end{subfigure}
    \vspace{-0.2em}
    \caption{\textbf{Coverage of the top-$\mathbf{m}$ selected memory keys.} In each panel, the solid line is the layer-averaged mean cumulative mass curve across all tokens; the shaded band shows the corresponding $[p10, p90]$ range. Panel (a) shows write keys and panel (b) shows read keys. Write keys exhibit a less uniform distribution than read keys.}
    \label{fig:topk-cumulative}
\end{figure}

\paragraph{Adaptive Memory Access Patterns.}
The memory slot utilization is controlled by the softmax over read and write keys. We analyze the peakiness of these distributions to assess whether all keys are actively used.
We pre-train SWA:SDM 1.4B models with varying read/write configurations ($W \in \{32,64\}$, $R \in \{32,64,128\}$) and extract softmax values per token across all layers and heads (for details see Appendix~\ref{app:cum_mass}).
First, we compute cumulative mass in the top-$m$ keys: $C(m)=\sum_{i=1}^{m} p_{(i)}$ where $p_{(1)} \geq p_{(2)} \geq \cdots \geq p_{(k)}$.
Figure~\ref{fig:topk-cumulative} shows that write distributions are moderately peaked: with $k=64$ writes, the top 32 keys capture $\sim$85\% of probability mass.
Read distributions are more uniform: with $k=128$ reads, the top 64 keys hold only $\sim$77\% of mass, indicating broader access patterns.
Second, we compute the effective number of keys per token as $u(p)=\exp(\mathrm{ent}(p))/k \in (0,1]$ (Appendix~\ref{app:cum_mass}), which measures utilization independent of $k$.

\captionsetup{font=small}

\begin{figure}[t]
    \centering
    \begin{subfigure}[t]{0.48\textwidth}
        \centering
        \includegraphics[width=\linewidth]{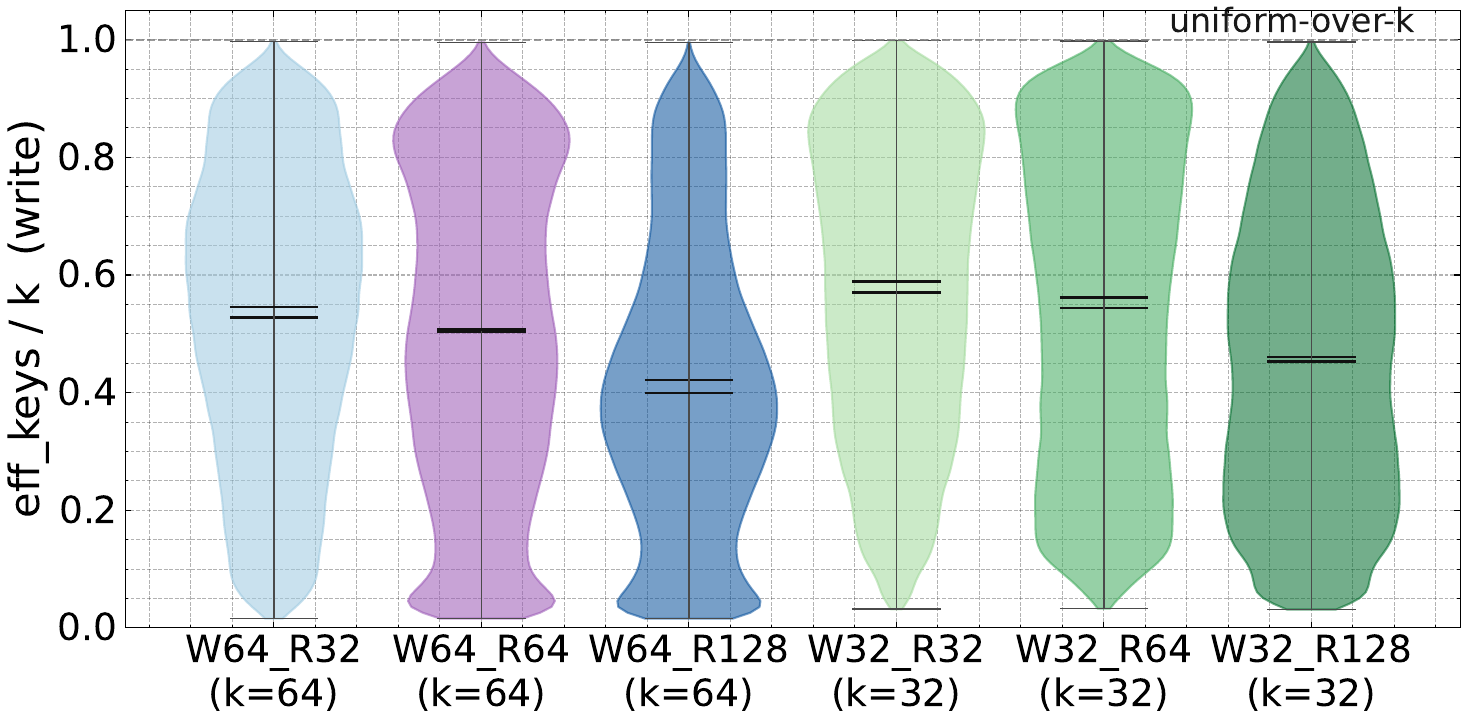}
        \caption{Write keys}
        \label{fig:effective-k-write}
    \end{subfigure}
    \hfill
    \begin{subfigure}[t]{0.48\textwidth}
        \centering
        \includegraphics[width=\linewidth]{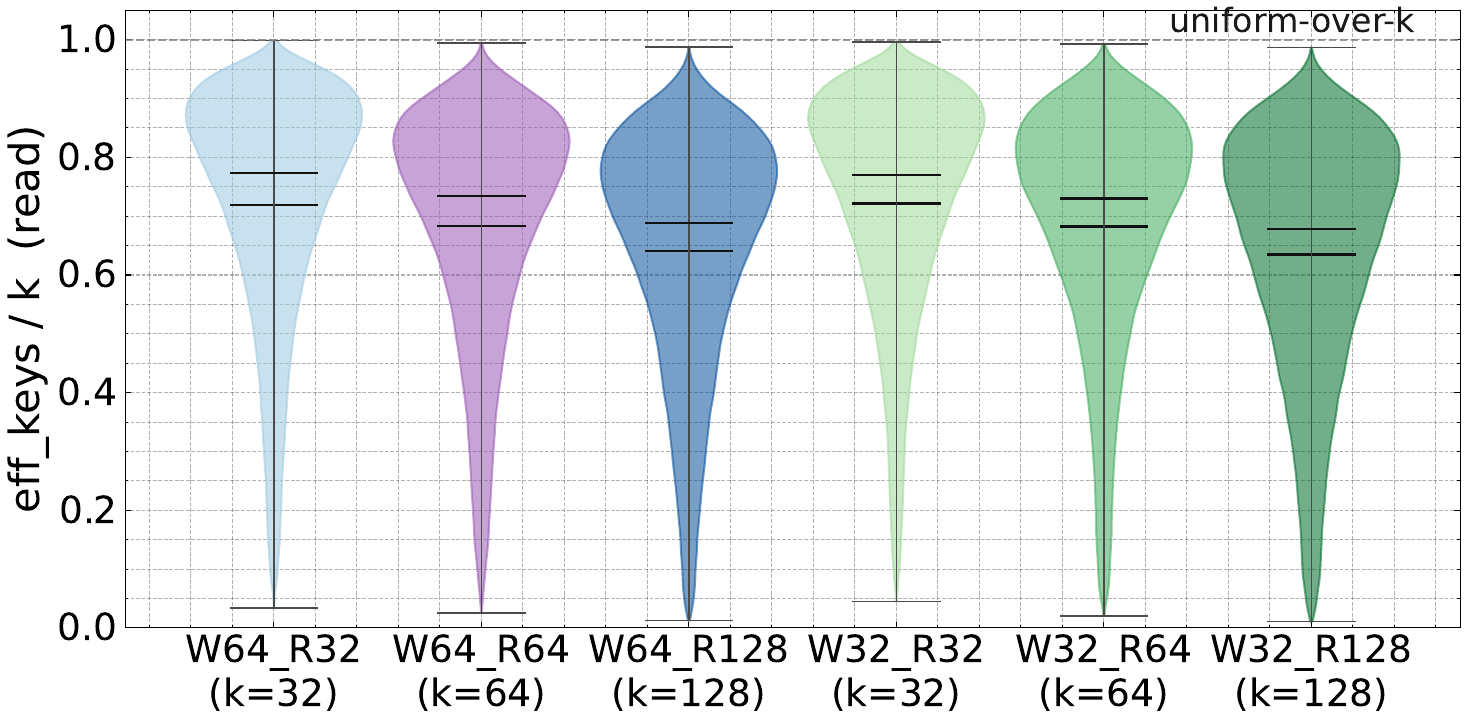}
        \caption{Read keys}
        \label{fig:effective-k-read}
    \end{subfigure}
    \caption{\textbf{Per-token effective-key utilization: analysis of $\mathbf{eff\_keys}/k$.} Each violin is built from all pooled token-level values for each run; the overlaid markers indicate the mean and median. Panel (a) shows write keys and panel (b) shows read keys. Read keys have a more uniform distribution than write keys.}
    \label{fig:effective-k}
\end{figure}

Figure~\ref{fig:effective-k} reveals distinct read/write behaviors: reads are consistently more uniform than writes across all configurations.
Increasing reads to 128 makes the distribution \textit{less} uniform (more selective), while decreasing writes to 32 makes writes \textit{more} uniform.
Notably, read and write distributions adapt to each other: with limited reads ($R=32$), writes become more uniform to compensate; with abundant reads ($R=128$), writes become more peaked.
This demonstrates SDM's adaptive memory access: the model dynamically balances read selectivity against write dispersion based on available capacity.
See Appendix~\ref{app:ablations} for additional details.

\captionsetup{font=normalsize}

\paragraph{Related Work.}
Fast-Weight Product Key Memory~\citep{zhao2026fastweightproductkeymemory} sparsifies Test-Time-Training methods~\citep{sun2025learninglearntesttime, zhang2025testtimetrainingright}
with a PKM memory. It differs from SDM in its update rule and in its hybridization and
ablation choices, is not evaluated under an isoFLOP constraint, and reports only
small-scale results, with at most 100M non-embedding parameters.

Sparse State Expansion (SSE)~\citep{pan2025scalinglinearattentionsparse} also shares our premise that long-context performance
is bounded by state size, and likewise enlarges the state by top-$k$ routing. The two designs differ in several ways. First, SSE  routes to whole partitions for efficiency, whereas SDM's product keys make slot-level selection tractable. 
Then, SDM applies the gated delta rule while SSE follows the gated linear attention (GLA) update rule~\citep{yang2024gatedlinearattentiontransformers} and thus has no error term.  
Moreover,
SDM routes reads and writes independently, whereas SSE always reads from the same partition set that it writes to.
Finally, SDM's addressing is trained by the language-modeling loss alone, whereas SSE
needs a load-balancing term to avoid collapse.

\section{Limitations}
While SDM strongly improves over previous linear RNNs, a gap to full attention remains on
tasks requiring multiple distinct retrievals from a long context, such as multikey 2 and
multikey 3 from the RULER benchmark. Targeted training partially mitigates this but there remains an intrinsic limitation to having a bounded state (Appendix~\ref{sec:app_collisions}).
Further work is therefore needed on architectures or frameworks which allow
fixed-state models to better deal with dense recall tasks \citep{zhang2026recursivelanguagemodels}.

\section{Conclusion}


We have introduced (i)~Sparse Delta Memory, a sparsification of the gated
delta rule whose product-key addressing scales the recurrent state by three orders of
magnitude at identical FLOPs; (ii)~a chunkwise-parallel training algorithm, built on
sparse interaction matrices and an in-place backward pass, that makes this regime
trainable at scale; (iii)~a learned initial state that additionally turns the memory into
parametric storage; and (iv)~an isoFLOP evaluation up to 8B parameters and 1T tokens,
where SDM improves over GDN in training loss, short-context accuracy and long-context
recall.
We believe that SDM opens up new possibilities for developing agents with improved long-term memory, and more generally for in-context learning over extended sequences such as video streams which are central to world modeling and robotics.

\bibliography{references}
\bibliographystyle{references}

\appendix

\pagebreak

\begin{center}
{\LARGE \bf Appendices}
\end{center}

\section{SDM Layer Diagram}
\label{sec:app_sdm_layer}

Figure~\ref{fig:sdm_architecture} details one SDM layer: the operations it shares with GDN and the ones SDM modifies. Figure~\ref{fig:gdn_to_sdm} in the main text gives the overview.

\begin{figure}[h]
\centering
\resizebox{0.9\textwidth}{!}{%
\begin{tikzpicture}[
    box/.style={rectangle, draw, rounded corners=2pt, font=\normalsize, thick, align=center},
    mem/.style={box, fill=violet!8, draw=violet!50, minimum width=1.2cm, minimum height=1.8cm, very thick},
    graybox/.style={box, fill=gray!8, draw=gray!50, font=\small, minimum width=0.9cm, minimum height=0.5cm, rounded corners=1pt},
    purplebox/.style={box, fill=violet!10, draw=violet!50, font=\footnotesize, minimum width=1.4cm, minimum height=0.7cm},
    purplevec/.style={box, fill=violet!10, draw=violet!50, font=\small, minimum width=0.9cm, minimum height=0.5cm, rounded corners=1pt},
    trapproj/.style={trapezium, trapezium left angle=75, trapezium right angle=75, draw, thick, fill=gray!8, draw=gray!50, font=\normalsize, minimum width=0.7cm, minimum height=0.45cm, trapezium stretches body},
    pkm/.style={box, fill=violet!10, draw=violet!50, font=\normalsize, minimum width=1.5cm, minimum height=1.1cm, densely dashed},
    post/.style={box, fill=gray!8, draw=gray!50, font=\normalsize, minimum width=1.1cm, minimum height=0.6cm},
    arr/.style={-stealth, thick, gray!70},
    purplearr/.style={-stealth, thick, violet!60},
]

\node[mem] (mt0) at (7, 2.8) {$\mathbf{M}_{t\text{-}1}$\\[-1pt]{\small $N\!\times\!d_v$}};
\draw[violet!50, thick, dashed, -stealth] ([xshift=-0.4cm]mt0.west) -- (mt0.west);
\node[circle, draw=violet!60, fill=violet!10, thick, densely dashed, minimum size=0.8cm, inner sep=0pt, font=\normalsize] (fgate) at (8.3, 2.8) {$\boldsymbol{\times}$};
\draw[violet!60, very thick] (mt0) -- (fgate);
\node[mem] (mtd) at (9.8, 2.8) {$\tilde{\mathbf{M}}_{t}$\\[-1pt]{\small $N\!\times\!d_v$}};
\draw[purplearr, very thick] (fgate) -- (mtd);
\node[ellipse, draw=violet!60, fill=violet!10, thick, densely dashed, inner sep=4pt, font=\small] (wadd) at (12.5, 2.8) {$+\; \beta_t \!\cdot\! k_t \!\otimes\! \Delta v_t$};
\draw[violet!60, very thick] (mtd) -- (wadd);
\node[mem] (mt1) at (16.0, 2.8) {$\mathbf{M}_{t}$\\[-1pt]{\small $N\!\times\!d_v$}};
\draw[purplearr, very thick] (wadd) -- (mt1);
\draw[violet!50, thick, -stealth, dashed] (mt1.east) -- (19.5, 2.8);


\node[purplebox, densely dashed] (read_k) at (9.8, 0.8) {$\tilde{\mathbf{M}}_t^\top k_t$};
\node[graybox] (vtilde) at (11.15, 0.8) {$\tilde{v}_t$};
\draw[purplearr] (read_k) -- (vtilde);
\draw[purplearr] (mtd.south) -- (read_k.north);

\node[graybox, font=\footnotesize, minimum width=1.4cm] (delta) at (12.5, 0.8) {$v_t - \tilde{v}_t$};
\node[graybox] (deltav_vec) at (12.5, 1.6) {$\Delta v_t$};

\node[purplebox, densely dashed] (read_q) at (16.0, 0.8) {$y_t = \mathbf{M}_t^\top q_t$};
\draw[purplearr] (mt1.south) -- (read_q.north);

\node[post] (ogate) at (18.2, 0.8) {$g_t \!\odot\! \text{LN}(y_t)$};
\draw[arr] (read_q) -- (ogate);
\node[post] (Wout) at (20, 0.8) {$W_o$};
\draw[arr] (ogate) -- (Wout);

\node[graybox] (ot) at (20.0, -0.3) {$o_t \!\in\!\mathbb{R}^d$};
\draw[arr] (Wout) -- (ot);


\node[graybox] (gamma_v) at (8.3, -0.3) {$\alpha_t \in\mathbb{R}$};
\node[purplevec, densely dashed] (ki_v) at (9.8, -0.3) {$k_t \!\in\!\mathbb{R}^N$};
\node[graybox] (v_v) at (12.5, -0.3) {$v_t \!\in\!\mathbb{R}^{d_v}$};
\node[graybox] (beta_v) at (14, -0.3) {$\beta_t \!\in\!\mathbb{R}$};
\node[purplevec, densely dashed] (qj_v) at (16.0, -0.3) {$q_t \!\in\!\mathbb{R}^N$};
\node[graybox] (og_v) at (18.2, -0.3) {$g_t \!\in\!\mathbb{R}^d$};

\draw[arr] (gamma_v) -- (fgate.south);

\draw[purplearr] (ki_v.north) -- (read_k.south);

\draw[arr] (delta) -- (deltav_vec);
\draw[arr] (vtilde.east) -- (delta.west);
\draw[arr] (v_v.north) -- (delta.south);
\draw[purplearr] (read_k) -- (vtilde);
\draw[purplearr] (mtd.south) -- (read_k.north);

\coordinate (merge) at (12.5, 2.1);
\draw[gray!70, thick] (deltav_vec.north) |- (merge);
\draw[gray!70, thick] (beta_v.north) |- (merge);
\draw[arr] (merge) -- (wadd.south);

\draw[purplearr] (qj_v.north) -- (read_q.south);

\draw[arr] (og_v) -- (ogate.south);

\node[pkm] (pkm_k) at (9.8, -1.45) {\small top-$W(k'_1\!\oplus\!k'_2)$\\[-2pt]\small $k'\!\to\![k'_1,k'_2]$};
\draw[purplearr] (pkm_k) -- (ki_v);

\node[pkm] (pkm_q) at (16.0, -1.45) {\small top-$R(q'_1\!\oplus\!q'_2)$\\[-2pt]\small $q'\!\to\![q'_1,q'_2]$};
\draw[purplearr] (pkm_q) -- (qj_v);

\node[trapproj] (Wa) at (8.3, -3.5) {$W_\alpha$};
\draw[arr] (Wa) -- (gamma_v);

\node[trapproj] (Wk) at (9.8, -3.5) {$W_k$};
\node[purplevec] (kprime) at (9.8, -2.6) {$k' \!\in\!\mathbb{R}^{2\sqrt{N}}$};
\draw[arr] (Wk) -- (kprime);
\draw[purplearr] (kprime) -- (pkm_k);

\node[trapproj] (Wv) at (12.5, -3.5) {$W_v$};
\draw[arr] (Wv) -- (v_v);

\node[trapproj] (Wb) at (14, -3.5) {$W_\beta$};
\draw[arr] (Wb) -- (beta_v);

\node[trapproj] (Wq) at (16.0, -3.5) {$W_q$};
\node[purplevec] (qprime) at (16.0, -2.6) {$q' \!\in\!\mathbb{R}^{2\sqrt{N}}$};
\draw[arr] (Wq) -- (qprime);
\draw[purplearr] (qprime) -- (pkm_q);

\node[trapproj] (Wog) at (18.2, -3.5) {$W_g$};
\draw[arr] (Wog) -- (og_v);

\node[graybox, minimum width=1.2cm, minimum height=0.5cm] (xt) at (7, -4.2) {$x_t \!\in\!\mathbb{R}^d$};
\draw[gray, thick] (xt.east) -- (18.2, -4.2);
\draw[gray, -stealth, thick] (8.3, -4.2) -- (Wa.south);
\draw[gray, -stealth, thick] (9.8, -4.2) -- (Wk.south);
\draw[gray, -stealth, thick] (12.5, -4.2) -- (Wv.south);
\draw[gray, -stealth, thick] (14, -4.2) -- (Wb.south);
\draw[gray, -stealth, thick] (16.0, -4.2) -- (Wq.south);
\draw[gray, -stealth, thick] (18.2, -4.2) -- (Wog.south);

\end{tikzpicture}
}
\caption{\textbf{SDM layer.} \textcolor{gray!60}{Gray}: operations present in GDN. \textcolor{violet!60}{Purple}: operations modified in SDM. Dashed borders indicate sparse operations ($W$ or $R$ out of $N$ slots). Symbols: $x_t$ layer input; $k'$, $q'$ pre-PKM key and query, split into halves; $k_t$, $q_t$ sparse write and read vectors; $v_t$ value; $\alpha_t$, $\beta_t$ forget and input gates (projections $W_\alpha$, $W_\beta$); $g_t$ output gate (projection $W_g$); $\mathbf{M}$ memory ($N$ slots of width $d_v$ per head); $y_t$ readout; LN per-head LayerNorm; $o_t$ layer output; $\oplus$ outer sum; $\otimes$ outer product; $\odot$ element-wise product.}
\label{fig:sdm_architecture}
\end{figure}

\section{Efficient Training of SDM}
\label{sec:efficient-training}

Training SDM requires computing exact outputs and gradients for the gated delta rule (Eq.~\ref{eq:sdm-decay}--\ref{eq:sdm-read}) across long sequences. We decompose this into \emph{intra-chunk parallel} computation (batched across all chunks) and \emph{chunkwise recurrent} computation (sequential across chunks), following the WY representation approach from GDN~\citep{yang2024parallelizinglineartransformersdeltan}.
Our kernels implement this WY formulation; see our code release (\url{https://github.com/facebookresearch/sparse-delta-memory}) for implementation details.

\subsection{Intra-Chunk Parallel vs.\ Chunkwise Recurrent}

We split the sequence into chunks of size $C$. Within each chunk, the gated delta rule creates causal dependencies between tokens: token $t$'s memory read may depend on earlier writes within the same chunk. The WY representation~\citep{yang2024parallelizinglineartransformersdeltan} resolves these dependencies analytically via a triangular solve, enabling parallel computation within each chunk.

\paragraph{Phase 1: Intra-chunk parallel (batched).}
All chunks can be processed in parallel. Let $\mathbf{V} \in \mathbb{R}^{C \times d_{\text{v}}}$ denote the stacked value vectors for a chunk, and $\boldsymbol{\beta} \in \mathbb{R}^C$ the per-token input gates. For each chunk, we compute:
\begin{itemize}
\item \textbf{Segmented cumulative decay.} For each slot $s$ accessed in the chunk, accumulate the log-decay $\log \alpha_t$ across tokens that write to slot $s$. This is a segmented prefix sum over slot indices: writing $(t, n)$ for the $n$-th of token $t$'s $W$ write entries, which addresses slot $s = \mathcal{I}^w_t[n]$, the cumulative log-decay is $\lambda_{t,n} = \sum_{t' \leq t, \, \mathcal{I}^w_{t'} \ni s} \log \alpha_{t'}$, summed over the tokens $t'$ of the current chunk.

\item \textbf{Sparse interaction matrices $\mathbf{G}$ and $\mathbf{QK}$.} We define two $C \times C$ lower-triangular matrices via the gated sparse inner product (Eq.~\ref{eq:sparse-ip}):
(i)~$\mathbf{G}[i,j]$, the \emph{write--write} self-interaction using write keys on both sides, capturing how token~$j$'s write to shared slots affects token~$i$'s delta-rule subtraction; and
(ii)~$\mathbf{QK}[i,j]$, the \emph{read--write} cross-interaction using read queries for token~$i$ and write keys for token~$j$, capturing how token~$j$'s write affects token~$i$'s read output.
Two tokens interact only if their selected slots overlap (the top-$W$ write slots; for $\mathbf{QK}$, the read slots of token~$i$), so with $W \ll N$ both matrices are typically very sparse. We compute them as \emph{sparse inner products} (Section~\ref{sec:sparse-ip}).

\item \textbf{Triangular solve.} Construct the unit lower-triangular matrix $\mathbf{M}_{\text{sys}} = \mathbf{I} + \text{diag}(\boldsymbol{\beta}) \, \mathbf{G}$ ($\mathbf{G}$ is strictly lower-triangular) and solve $\mathbf{M}_{\text{sys}} \, \Delta\mathbf{V}_{\text{const}} = \text{diag}(\boldsymbol{\beta}) \cdot \mathbf{V}$. This yields the memory-independent part of the intra-chunk delta-v values (their value if the chunk-initial memory were empty). We also define $\mathbf{B} = \mathbf{M}_{\text{sys}}^{-1} \cdot \text{diag}(-\boldsymbol{\beta})$, which maps reads from the chunk-initial memory to their contribution to the delta-v values.
\end{itemize}

None of the operations above depends on the memory state, so they can be batched over all chunks.

\paragraph{Phase 2: Chunkwise recurrent (sequential).}
Chunks are processed one at a time, since each chunk's memory writes depend on the previous chunk's memory state. Per chunk:
\begin{enumerate}[leftmargin=*,itemsep=2pt]
\item \textbf{Read:} Read the chunk-initial memory at write indices to obtain $\texttt{retrieved}[t] = \sum_{n} k_{\text{eff},t}^{(n)} \cdot \mathbf{M}[\mathcal{I}^w_t[n]]$ (where $k_{\text{eff}} = k_t^{(i)} \cdot e^{\lambda}$ and $q_{\text{eff}} = q_t^{(i)} \cdot e^{\lambda^r}$ incorporate cumulative decay, for the selected key values $k_t^{(i)}$ and read weights $q_t^{(i)}$), and at read indices to obtain the inter-chunk output $\texttt{inter}[t] = \sum_{n} q_{\text{eff},t}^{(n)} \cdot \mathbf{M}[\mathcal{I}^r_t[n]]$.
\item \textbf{Correct:} Compute the full delta-v incorporating the current memory state: $\Delta\mathbf{V} = \Delta\mathbf{V}_{\text{const}} + \mathbf{B} \cdot \texttt{retrieved}$. The term $\mathbf{B} \cdot \texttt{retrieved}$ adds the contribution of the chunk-initial memory state, propagated through the intra-chunk dependencies by $\mathbf{M}_{\text{sys}}^{-1}$. Row $t$ of $\Delta\mathbf{V}$, $\beta_t \Delta\mathbf{v}_t = \beta_t (\mathbf{v}_t - \mathbf{\Tilde{v}}_t)$, is the residual written in Eq.~\eqref{eq:sdm-write}, with the input gate folded in and the intra-chunk dependence of $\mathbf{\Tilde{v}}_t$ on earlier writes resolved by the triangular solve rather than sequentially.
\item \textbf{Output:} Compute the final chunk output combining inter-chunk reads with intra-chunk corrections: $\mathbf{y} = \texttt{inter} + \text{tril}(\mathbf{QK}) \cdot \Delta\mathbf{V}$.
\item \textbf{Write:} Apply the per-slot decay to the written slots and add the delta-v contributions (Eq.~\eqref{eq:sdm-write}).
\end{enumerate}

\subsection{Sparse Inner Product}
\label{sec:sparse-ip}

The interaction matrices $\mathbf{G}[i,j]$ and $\mathbf{QK}[i,j]$ measure how much tokens $i$ and $j$ interact through shared memory slots. For GDN with dense keys, these are decay-weighted, causally masked dense products (built from $\mathbf{K}\mathbf{K}^\top$ and $\mathbf{Q}\mathbf{K}^\top$, where $\mathbf{K}, \mathbf{Q} \in \mathbb{R}^{C \times d_{\text{qk}}}$ stack the dense keys and queries of a chunk). For SDM with sparse keys, two tokens interact only if their selected slots overlap, so most entries are typically zero.

We compute the \emph{gated sparse inner product}:
\begin{equation}
\mathbf{G}[i,j] = \sum_{\substack{n,m \,:\, \mathcal{I}^w_i[n] = \mathcal{I}^w_j[m]}} k_i^{(n)} \cdot k_j^{(m)} \cdot \exp\!\left(\lambda_{i,n} - \lambda_{j,m}\right), \quad j < i
\label{eq:sparse-ip}
\end{equation}
where $k_i^{(n)}$ is the key value of the $n$-th selected slot $\mathcal{I}^w_i[n]$, the sum ranges only over matching slot indices between tokens $i$ and $j$, and $\lambda$ are the cumulative log-decays.
$\mathbf{QK}[i,j]$ is defined identically, except that token $i$ enters through its \emph{read} set instead of its write set:
\begin{equation}
\mathbf{QK}[i,j] = \sum_{\substack{n,m \,:\, \mathcal{I}^r_i[n] = \mathcal{I}^w_j[m]}} q_i^{(n)} \cdot k_j^{(m)} \cdot \exp\!\left(\lambda^r_{i,n} - \lambda_{j,m}\right), \quad j \leq i
\label{eq:sparse-ip-qk}
\end{equation}
where $q_i^{(n)}$ is the read weight of token $i$ on slot $\mathcal{I}^r_i[n]$ and $\lambda^r_{i,n}$ is the same cumulative log-decay evaluated at that slot and position. The diagonal $j = i$ is included, since a token reads the memory after its own write (Eq.~\eqref{eq:sdm-read}), whereas the delta-rule subtraction captured by $\mathbf{G}$ is strictly causal.

A dense inner product (GDN) sums $O(C^2 \cdot d_{\text{qk}})$ terms per chunk, whereas the sparse version involves at most $O(C^2 \cdot W)$ nonzero terms, where $W\,{=}\,64 \ll N$, independent of the number of  slots $N$.

\subsection{Memory-Efficient Backward}

Following Sparse Access Memory~\citep{rae2016scalingmemoryaugmentedneuralnetworks}, we apply sparse updates to the memory \textit{in-place} during the forward pass and \emph{undo} these sparse updates during the backward pass to recover the correct memory state for each timestep. This avoids storing a full copy of the $N \times d_{\text{v}}$ memory at each chunk boundary, reducing peak memory from $O((T/C) \times N \times d_{\text{v}})$ to $O(N \times d_{\text{v}} + T \times W \times d_{\text{v}})$ for sequence length $T$. For example, at a 16384-token training sequence length and for an 8B training with chunk size $C=128$ and $W=64$, this reduces the memory consumed by the memory-state checkpoints of an SDM layer from 226\,GB to 8\,GB.
For longer sequence lengths, such as in 128k long-context finetuning, this memory snapshot can become prohibitively large.
We therefore investigated quantization of the snapshot to fp8 and int4, which makes the recovered memory state approximate, and found that both quantization levels seem to have no detrimental effect on training loss or end performance; we therefore recommend it for long-context finetuning.

\ifmultifactor
\section{Multi-Factor Product Keys}
\label{sec:app_multifactor}

The product key of Section~\ref{sec:sdm-method} has two factors: it is the
direct application of PKM~\citep{lample2019largememorylayersproduct}. The rest of SDM does not depend
on this choice. This appendix generalizes the addressing to $F$ factors, explains how the exact top-$k$ is
computed, and reports what the number of factors changes at 1.4B scale.

\subsection{\texorpdfstring{From Two to $F$ Factors}{From Two to F Factors}}

For each head, the write-key projection outputs a vector $\mathbf{k}'_t\in\mathbb{R}^{d_{\text{qk}}}$, which we
split into $F$ sub-vectors $\mathbf{k}'_{f,t}\in\mathbb{R}^{n_f}$, $f=1,\dots,F$, with
$d_{\text{qk}}=\sum_f n_f$. A slot is a tuple $c=(c_1,\dots,c_F)$ with $c_f\in\{0,\dots,n_f-1\}$, so the head
addresses $N=\prod_f n_f$ slots, and the score of slot $c$ is the $F$-way outer sum
\begin{equation}
s_t(c)=\sum_{f=1}^{F}\mathbf{k}'_{f,t}[c_f],
\qquad\text{i.e.}\qquad
\mathbf{s}_t=\mathbf{k}'_{1,t}\oplus\mathbf{k}'_{2,t}\oplus\cdots\oplus\mathbf{k}'_{F,t}\in\mathbb{R}^{n_1\times\cdots\times n_F}.
\label{eq:multifactor-score}
\end{equation}
Read scores are built in the same way from $\mathbf{q}'_t$. Slot $c$ is row
$(\cdots((c_1 n_2+c_2)\,n_3+c_3)\cdots)\,n_F+c_F$ of the head's memory table. The factorization changes
the addressing scores, hence the selected slots and their normalized key values; the top-$W$ and top-$R$
selections, the gated delta write, the sparse read and the chunkwise training algorithm
(Appendix~\ref{sec:efficient-training}) keep the same form. With $F=2$ and $n_1=n_2=\sqrt{N}$, this is
the product key of Section~\ref{sec:sdm-method}. Figure~\ref{fig:mf_tesseract} shows the slots selected in one head of an 8B
model with four factors of 16.

\paragraph{Slots versus addressing parameters.}
The key and query projections of a head have $d_{\text{qk}}=\sum_f n_f$ output columns each, while the head
addresses $N=\prod_f n_f$ slots. For fixed $d_{\text{qk}}$ and $F$, $N\le(d_{\text{qk}}/F)^F$ (AM--GM); integer factors that
differ by at most one maximize $N$, with equality when $F$ divides $d_{\text{qk}}$, and the bound grows with $F$ as
long as $F<d_{\text{qk}}/e$. Conversely, balanced factors address $N$ slots with $d_{\text{qk}}\approx F\,N^{1/F}$ columns. At
the 57{,}600 slots per head of the 1.4B model, two factors $[240,240]$ need $d_{\text{qk}}=480$, four factors
$[15,16,16,15]$ need $d_{\text{qk}}=62$, and eight factors $[5,5,4,4,4,4,3,3]$ need $d_{\text{qk}}=32$. 
Summed over heads, the key and query projections have $Hd_{\text{qk}}$ columns each; at 8B this is 1{,}920, the $Hd_{\text{qk}}$ of the 30-head GDN baseline.
With two factors, keeping GDN's 30 heads gives $d_{\text{qk}}=64$ per head, factors $[32,32]$ and $N=1{,}024$ slots per head: a state expansion factor $N/d_{\text{qk}}$ of only 16 over GDN, whose total value width is the same.
This is why the two-factor SDM of the main experiments spends the same columns on fewer heads: at 8B, 2 heads with $d_{\text{qk}}=960$, i.e.\ $[480,480]$, an expansion factor of 3{,}600.
Four factors remove this trade-off: $[16,16,16,16]$ at $d_{\text{qk}}=64$ keeps the 30 heads of the GDN and full-attention baselines with an expansion factor of 1{,}024.
Tables~\ref{tab:mf_summary} and~\ref{tab:mf_heads} compare, at near-iso active parameters and state, two-factor models with few heads and four-factor models with many heads: the four-factor models score higher on RULER.  

\begin{figure}[t]
\centering
\includegraphics[width=0.8\linewidth]{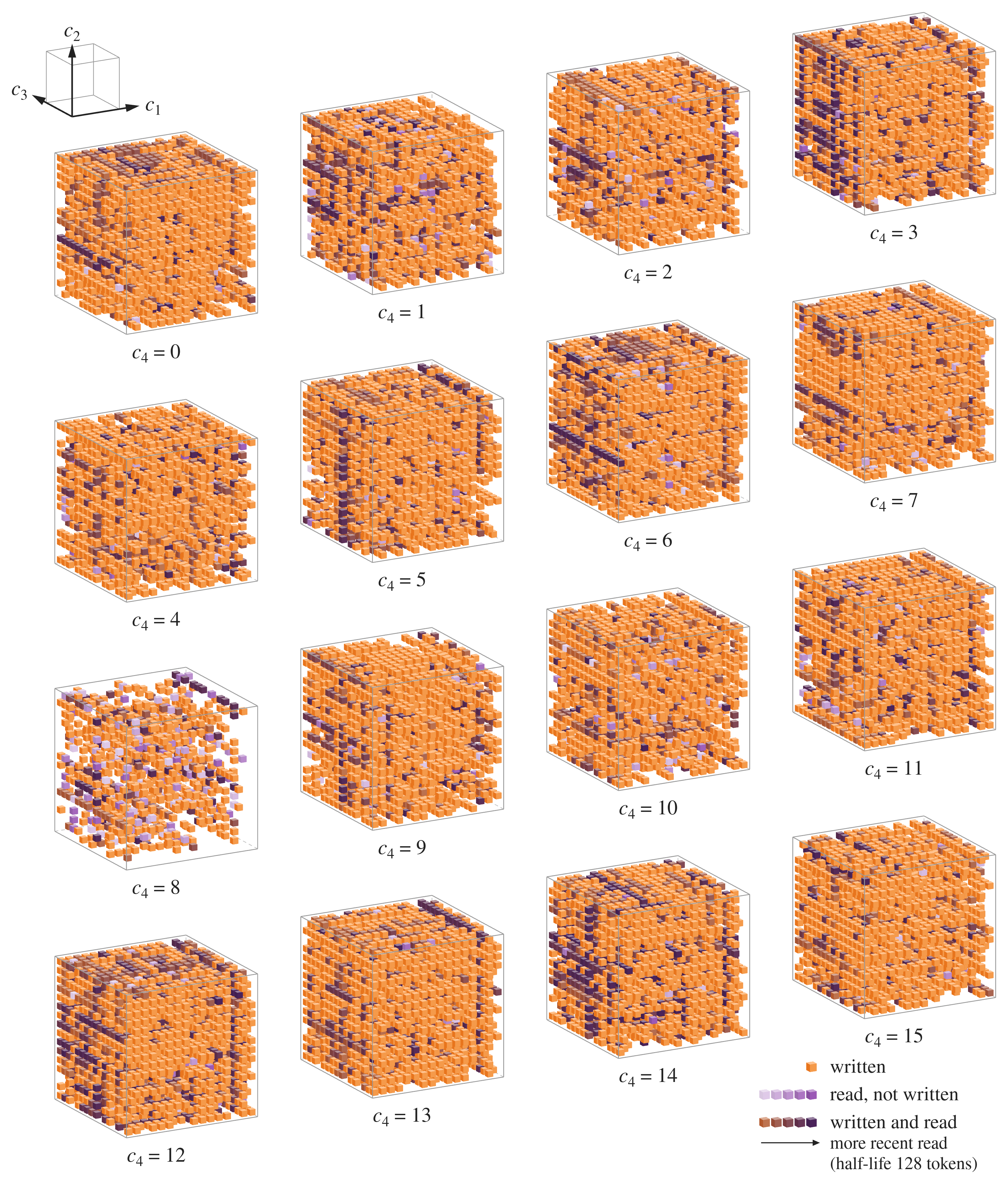}
\caption{\textbf{One SDM head's memory with four factors} (8B model with 30 SDM heads per layer, factor
dimensions $[16,16,16,16]$, $N=65{,}536$ slots per head; layer 7, head 11). Slot $(c_1,c_2,c_3,c_4)$ is
drawn in the cube of its value of $c_4$, at position $(c_1,c_2,c_3)$. After one 8{,}192-token sequence,
65.8\% of the slots have been selected by at least one top-64 write (orange). Purple marks the slots read
in the last 723 tokens, darker for more recent reads (a display choice: half-life of 128 tokens), and dark
purple if they were also written. The slots written for one token are unevenly concentrated across the values of
each factor, which shows as lines and slabs. Some factor values are rarely selected in this sequence, e.g.\ 20\% of the slots with
$c_4\,{=}\,8$ are written.}
\label{fig:mf_tesseract}
\end{figure}

\subsection{\texorpdfstring{Exact Top-$k$ over an $F$-Way Outer Sum}{Exact Top-k over an F-Way Outer Sum}}

As in PKM, the top-$k$ of an outer sum involves only the top-$k$ of each operand:
$\text{top}_k(\mathbf{a}\oplus\mathbf{b})=\text{top}_k(\text{top}_k(\mathbf{a})\oplus\text{top}_k(\mathbf{b}))$.
Since $\oplus$ is associative, we keep the $\min(k,n_f)$ largest entries of each factor and merge the
factors one at a time, keeping the best $\min(k,\ell\ell')$ entries of each merge of two lists of lengths
$\ell$ and $\ell'$ (the running merged list, $\ell\le k$, and the next factor's $\min(k,n_f)$ largest entries). After the $F-1$ merges, this gives the exact top-$k$ of Eq.~\eqref{eq:multifactor-score},
with $k=W$ for writes and $k=R$ for reads (exact ties may be broken either way).

Each merge combines two lists sorted in decreasing order, of lengths $\ell,\ell'\le k$. It does not need all
$\ell\ell'$ sums: a pair of ranks $(i,j)$ (0-based) with $(i+1)(j+1)>k$ is dominated by the
$(i+1)(j+1)-1\ge k$ pairs of lower or equal ranks, which all score at least as much, so only the
pairs with $(i+1)(j+1)\le k$ are scored. There are at most $\sum_{a=1}^{k}\lfloor k/a\rfloor\le
k(1+\ln k)$ of them, 280 instead of 4{,}096 for $k=64$. With a linear-time selection among these
candidates, selection costs $O(d_{\text{qk}}\log k)$ per head and token for the per-factor top-$k$ and
$O(F\,k\log k)$ for the merges and for sorting the selected indices. Computing the factor scores costs $O(d\,d_{\text{qk}})$, the key projection
itself. None of these steps materializes the $N$ scores: they depend on $N$ only through $d_{\text{qk}}$, which is
about $F\,N^{1/F}$ for balanced factors. Our kernel fuses the selection and returns the slot indices in
increasing order, with their scores permuted accordingly; it adds the factors in order, rounding partial
sums to bfloat16 for bfloat16 inputs with more than two factors, and selects an exact top-$k$ under that
rounding. In the backward pass, the gradient of each selected score flows to its $F$ factor entries.

\paragraph{Sharing factors across groups of heads.}
Factors need not all be per head: a factor can be shared within each of $G$ groups of heads. With a
per-head factor of size $n_1$ and a factor of size $n_2$ shared within each group, the key projection has
$Hn_1+Gn_2$ output columns ($H(n_1+n_1^{\prime})+Gn_2$ with a per-head pair $n_1,n_1^{\prime}$). All heads of a group see
the same shared factor, so its $\min(k,n_2)$ largest entries are computed once per group and copied to
the heads before the merges.

\subsection{Experiments}
\label{sec:app_mf_experiments}

Each paragraph below makes one comparison, and the table beside it lists only the configurations
compared. Unless stated otherwise, all runs use the 1.4B (L08) SWA:SDM model of Table~\ref{tab:ladder}
(five SDM layers, $W=R=64$, value width $d_{\text{v}}=1920/H$ per head), pretrained for 80{,}320 steps
on 8k-token sequences (168.4B tokens) and fine-tuned for 1{,}000 steps on 128k-token sequences with the
same data and schedule. Only the number of SDM heads $H$ and the factor dimensions differ, and runs of
the same table share every other hyperparameter, except where a footnote says otherwise. 

\paragraph{Overview.}
Table~\ref{tab:mf_summary} sets a two-factor and a four-factor model side by side at each scale, in the
format of Table~\ref{tab:l08_accuracy}; the two models of each pair share their data and training schedule. With four factors, the RULER average over the 13
tasks is higher at both scales (by 6.5 points at 1.4B and 2.8 points at 8B), and at 1.4B the
short-context tasks are level. The largest per-task decline is in variable tracking (vt), which is lower
with four factors at both scales. We have observed vt to be a noisy task: GDN's vt accuracy even increases with sequence length
(Figure~\ref{fig:ruler_per_task_1.4b}), an artefact of the task design. We therefore give more weight to the NIAH tasks
than to this single vt result.

\begin{table}[!htbp]
\centering
{\small
\caption{\textbf{Two and four factors, 1.4B and 8B parameter scales} (after long-context
fine-tuning). 1.4B: the two-head $[240,240]$ and 16-head $[15,16,16,15]$ runs of Table~\ref{tab:mf_sweep}
(same state; same data, steps and batch); GDN: the model of
Table~\ref{tab:l08_accuracy} (different recipe). 8B: four-head [240,240] and 30-head [16,16,16,16] models (not the two-head SDM of
Table~\ref{tab:l08_accuracy}); GDN: the model of Table~\ref{tab:l08_accuracy}. Metrics, MMLU included, as in
Table~\ref{tab:l08_accuracy}. PPL by position: perplexity over all tokens at positions $[16\text{k},32\text{k})$ and $[64\text{k},128\text{k})$ of the 2{,}305 code documents of the Metrics paragraph (Appendix~\ref{sec:app_multifactor}), each read from its start. Subscripts: SDM 4F minus SDM 2F, in green when SDM 4F is better or equal
and in red when it is worse (lower NLL and PPL are better).
\label{tab:mf_summary}}
\smallskip
\begin{tabular}{@{\ }p{3.7cm}|rrr|rrr@{\ }}
\toprule
& \multicolumn{3}{c|}{1.4B (L08)} & \multicolumn{3}{c}{8B (L13)} \\
\multicolumn{1}{r|}{Local layer $\rightarrow$} & \multicolumn{3}{c|}{SWA} & \multicolumn{3}{c}{SWA} \\
\multicolumn{1}{r|}{Global layer $\rightarrow$} & GDN & SDM 2F & SDM 4F\quad~ & GDN & SDM 2F & SDM 4F\quad~ \\
\multicolumn{1}{r|}{Heads $H$ $\rightarrow$} & 15 & 2 & 16\quad~ & 30 & 4 & 30\quad~ \\
\multicolumn{1}{r|}{State (total) $\rightarrow$} & 614k & 553M & 553M\quad~ & 2.2M & 1.99B & 2.26B\quad~ \\
\midrule
Validation text NLL {\small$\downarrow$} & 2.683 & 2.667 & \textbf{2.663} \subg{-0.004} & 2.298 & 2.289 & \textbf{2.288} \subg{-0.001} \\
\midrule
PPL, positions $[16\text{k},32\text{k})$ {\small$\downarrow$} & 2.270 & 2.099 & \textbf{2.057} \subg{-0.042} & 1.815 & 1.752 & \textbf{1.742} \subg{-0.010} \\
PPL, positions $[64\text{k},128\text{k})$ {\small$\downarrow$} & 2.237 & 2.028 & \textbf{1.965} \subg{-0.063} & 1.776 & 1.673 & \textbf{1.657} \subg{-0.016} \\
\midrule
HellaSWAG {\small$\uparrow$} & 61.27 & \textbf{62.19} & 61.01 \subr{-1.18} & 79.10 & \textbf{79.49} & 79.47 \subr{-0.02} \\
WinoGrande {\small$\uparrow$} & \textbf{61.56} & 60.22 & 60.69 \subg{+0.47} & 73.24 & 72.61 & \textbf{73.48} \subg{+0.87} \\
ARC easy {\small$\uparrow$} & 62.49 & 62.11 & \textbf{64.69} \subg{+2.58} & \textbf{79.15} & 78.48 & 77.72 \subr{-0.76} \\
ARC challenge {\small$\uparrow$} & 34.59 & \textbf{35.36} & 33.91 \subr{-1.45} & 52.27 & \textbf{52.45} & 50.82 \subr{-1.63} \\
PIQA {\small$\uparrow$} & 73.83 & \textbf{75.08} & 74.76 \subr{-0.32} & \textbf{81.01} & 80.03 & 79.76 \subr{-0.27} \\
OpenBookQA {\small$\uparrow$} & 36.80 & \textbf{38.00} & 36.60 \subr{-1.40} & 43.00 & \textbf{45.80} & 44.40 \subr{-1.40} \\
RACE.mid {\small$\uparrow$} & 50.42 & 50.00 & \textbf{50.70} \subg{+0.70} & 60.17 & \textbf{61.49} & 61.21 \subr{-0.28} \\
RACE.high {\small$\uparrow$} & 36.68 & 37.36 & \textbf{37.76} \subg{+0.40} & 43.40 & \textbf{44.68} & 44.63 \subr{-0.05} \\
CommonsenseQA {\small$\uparrow$} & 19.33 & \textbf{20.48} & 20.07 \subr{-0.41} & 66.83 & 63.47 & \textbf{69.70} \subg{+6.23} \\
BoolQ {\small$\uparrow$} & 63.03 & \textbf{63.33} & 62.81 \subr{-0.52} & 74.98 & \textbf{75.41} & 71.77 \subr{-3.64} \\
TQA {\small$\uparrow$} & 23.68 & 25.80 & \textbf{26.06} \subg{+0.26} & 55.36 & 56.72 & \textbf{57.01} \subg{+0.29} \\
HumanEval+/pass@1 {\small$\uparrow$} & \textbf{8.54} & 7.93 & \textbf{8.54} \subg{+0.61} & 18.29 & 20.73 & \textbf{21.95} \subg{+1.22} \\
NaturalQuestions {\small$\uparrow$} & 7.76 & 8.31 & \textbf{8.53} \subg{+0.22} & 22.60 & \textbf{23.96} & 23.74 \subr{-0.22} \\
MMLU {\small$\uparrow$} & \textbf{26.02} & 25.32 & 25.21 \subr{-0.11} & \textbf{57.24} & 56.79 & 55.88 \subr{-0.91} \\
GSM8K {\small$\uparrow$} & \textbf{2.96} & 1.90 & 2.27 \subg{+0.37} & 28.81 & \textbf{31.01} & 28.66 \subr{-2.35} \\
\midrule
Average accuracy {\small$\uparrow$} & 37.93 & 38.23 & \textbf{38.24} \subg{+0.01} & 55.70 & \textbf{56.21} & 56.01 \subr{-0.20} \\
\midrule
\multicolumn{7}{@{\ }l}{\textit{RULER long-context tasks (exact match, averaged over lengths 4k--131k)}} \\
single 1 {\small$\uparrow$} & 99.9 & 96.8 & \textbf{100.0} \subg{\sz+3.2} & \textbf{100.0} & \textbf{100.0} & \textbf{100.0} \subg{\sz+0.0} \\
single 2 {\small$\uparrow$} & 20.7 & 37.1 & \textbf{72.2} \subg{+35.1} & 45.1 & 87.7 & \textbf{89.2} \subg{\sz+1.5} \\
single 3 {\small$\uparrow$} & 12.1 & 34.6 & \textbf{53.6} \subg{+19.0} & 32.6 & 62.6 & \textbf{93.2} \subg{+30.6} \\
multikey 1 {\small$\uparrow$} & 13.6 & 24.2 & \textbf{32.4} \subg{\sz+8.2} & 32.9 & \textbf{65.7} & 64.5 \subr{\sz-1.2} \\
multikey 2 {\small$\uparrow$} & \sz0.7 & \sz0.6 & \textbf{\sz0.8} \subg{\sz+0.2} & \sz1.0 & \textbf{16.0} & 14.4 \subr{\sz-1.6} \\
multikey 3 {\small$\uparrow$} & \sz0.1 & \sz0.1 & \textbf{\sz0.6} \subg{\sz+0.5} & \sz0.1 & \sz2.2 & \textbf{\sz3.9} \subg{\sz+1.7} \\
multivalue {\small$\uparrow$} & 11.7 & 27.5 & \textbf{51.5} \subg{+24.0} & 31.1 & 71.7 & \textbf{78.0} \subg{\sz+6.3} \\
multiquery {\small$\uparrow$} & 11.8 & 31.4 & \textbf{52.5} \subg{+21.1} & 31.2 & 74.0 & \textbf{81.9} \subg{\sz+7.9} \\
vt {\small$\uparrow$} & 16.6 & \textbf{36.0} & 11.8 \subr{-24.2} & 46.2 & \textbf{80.4} & 55.9 \subr{-24.5} \\
cwe {\small$\uparrow$} & \sz6.6 & \sz4.3 & \textbf{\sz9.9} \subg{\sz+5.6} & 11.4 & \textbf{22.6} & 20.3 \subr{\sz-2.3} \\
fwe {\small$\uparrow$} & 35.6 & \textbf{49.9} & 45.1 \subr{\sz-4.8} & 62.7 & 62.8 & \textbf{75.9} \subg{+13.1} \\
qa 1 {\small$\uparrow$} & 13.3 & \textbf{14.3} & 14.0 \subr{\sz-0.3} & 23.6 & 26.5 & \textbf{29.9} \subg{\sz+3.4} \\
qa 2 {\small$\uparrow$} & 18.3 & \textbf{20.0} & 17.5 \subr{\sz-2.5} & 28.2 & 28.0 & \textbf{30.3} \subg{\sz+2.3} \\
\midrule
RULER average {\small$\uparrow$} & 20.1 & 29.0 & \textbf{35.5} \subg{\sz+6.5} & 34.3 & 53.9 & \textbf{56.7} \subg{\sz+2.8} \\
\bottomrule
\end{tabular}}
\end{table}

\paragraph{Metrics.}
Tables~\ref{tab:mf_heads}--\ref{tab:mf_factors} and Figure~\ref{fig:mf_curves} use the following
protocol. \emph{PPL} is computed on the last 1{,}024 tokens of the first $2^{20}$ tokens of each of 2{,}305 code
documents, as the exponential of the mean NLL over these target tokens. For $T\le512$k, the model sees
only the $T$ tokens before the targets, without a BOS token; $T=1$M denotes the whole window, which
starts with BOS (1{,}047{,}552 tokens before the targets). \emph{SLPL} (SlidingLongPPL) restricts the perplexity to key
tokens, following LongPPL~\citep{fang2025wrongperplexitylongcontextlanguage}: tokens to which an oracle
model (Llama-3.1-8B~\citep{grattafiori2024llama3herdmodels}) assigns a log-probability above $-2$ with the full context, and which gain more than 2
nats over a 4{,}096-token sliding window that keeps the BOS token. 
SLPL can be thought of as a fast approximation of LongPPL.
We report it for the tokens of the same 2{,}305 documents whose
position lies in $(T/2, T]$. In the tables that follow, bold marks the best value of a column, and the
coloured subscripts give the difference from the reference row, as in Table~\ref{tab:l08_accuracy}.

\paragraph{More heads at a constant state.}
With four factors, 16 heads fit in key projections only 3\% wider than those of the two heads used in
the main text (Table~\ref{tab:mf_heads}): the state is the same (eight times as many slots, each eight
times narrower), as are the per-token FLOPs of the memory reads and writes. The 16-head configuration has lower PPL at all
11 evaluated context lengths from 1k to 1M tokens (2.011 against 2.073 with 1M tokens of context), and
lower SLPL in every position bin from 8k to 128k.

\begin{table}[!htbp]
\centering\footnotesize
\setlength{\tabcolsep}{3.2pt}
\caption{\textbf{More heads at a constant state} (1.4B, protocol of the Metrics paragraph). $N$: slots per head; Key cols: output columns of each of
the key and query projections ($Hd_{\text{qk}}$ without shared factors); State: entries per SDM layer. Reference row: the configuration of the
main text. NIAH: mean exact match over the 8 needle-in-a-haystack tasks of RULER at 4k--128k (as in
Tables~\ref{tab:l08_accuracy} and~\ref{tab:mf_summary}), after the 128k fine-tuning.}
\label{tab:mf_heads}
\smallskip
\begin{tabular}{@{}rlrrr|ccc|cc|c@{}}
\toprule
 & & & & & \multicolumn{3}{c|}{PPL, context $T$ $\downarrow$} & \multicolumn{2}{c|}{SLPL, $T$ $\downarrow$} & \\
$H$ & Factor dims & $N$ & Key cols & State & 32k & 128k & 1M & 32k & 128k & NIAH $\uparrow$ \\
\midrule
2 & $[240,240]$ & 57{,}600 & 960 & 110.6M & 2.073 & 2.046 & 2.073 & 4.19 & 7.03 & 36.3 \\
16 & $[15,16,16,15]$ & 57{,}600 & 992 & 110.6M & \textbf{2.037}\subg{-0.036} & \textbf{1.990}\subg{-0.056} & \textbf{2.011}\subg{-0.062} & \textbf{3.47}\subg{-0.72} & \textbf{4.91}\subg{-2.12} & \textbf{40.9}\subg{+4.6} \\
\bottomrule
\end{tabular}
\end{table}
\FloatBarrier

\paragraph{A larger state at a similar parameter count.}
With the same 16 heads and the same 992 key columns, four factors ($[15,16,16,15]$) address 57{,}600 slots per
head, 60$\times$ more than the 961 slots of two factors ($[31,31]$); everything else is unchanged, including the
parameter count excluding the learned initial state $\mathbf{M}_0$ (Table~\ref{tab:mf_state}). The larger
state that four factors enable lowers PPL at every context length (2.011 against 2.174 with 1M tokens of
context) and reduces SLPL from 9.97 at 32k sequence length to 3.47. Moreover, the larger state size also raises NIAH from 27.2 to 40.9. Therefore, even at a fixed number of heads and addressing parameters,
four factors and the larger state they allow are beneficial.

\begin{table}[!htbp]
\centering\footnotesize
\setlength{\tabcolsep}{3.2pt}
\caption{\textbf{A larger state at a similar parameter count} (1.4B). Columns as in
Table~\ref{tab:mf_heads}. Reference row: two factors.}
\label{tab:mf_state}
\smallskip
\begin{tabular}{@{}rlrrr|ccc|cc|c@{}}
\toprule
 & & & & & \multicolumn{3}{c|}{PPL, context $T$ $\downarrow$} & \multicolumn{2}{c|}{SLPL, $T$ $\downarrow$} & \\
$H$ & Factor dims & $N$ & Key cols & State & 32k & 128k & 1M & 32k & 128k & NIAH $\uparrow$ \\
\midrule
16 & $[31,31]$ & 961 & 992 & 1.8M & 2.178 & 2.171 & 2.174 & 9.97 & 19.70 & 27.2 \\
16 & $[15,16,16,15]$ & 57{,}600 & 992 & 110.6M & \textbf{2.037}\subg{-0.141} & \textbf{1.990}\subg{-0.181} & \textbf{2.011}\subg{-0.163} & \textbf{3.47}\subg{-6.50} & \textbf{4.91}\subg{-14.79} & \textbf{40.9}\subg{+13.7} \\
\bottomrule
\end{tabular}
\end{table}
\FloatBarrier

\paragraph{Factors at a constant state.}
Among three runs with 16 heads and 57{,}600 slots per head, more factors mean fewer addressing
parameters (Table~\ref{tab:mf_factors}). Two factors use $7.7\times$ the key columns per head of four
factors (480 against 62, i.e.\ 128.5M more weights and biases over the five SDM layers) and give the
lowest PPL at all 11 evaluated context lengths, and the lowest pretraining loss; eight factors
give the highest PPL. SLPL orders the runs differently: four factors have the lowest SLPL in the 32k, 64k and 128k
position bins, and two factors in the 8k and 16k bins.
Figure~\ref{fig:mf_curves} shows all five runs of
Tables~\ref{tab:mf_heads}--\ref{tab:mf_factors} at every context length and position bin.

\begin{table}[!htbp]
\centering\footnotesize
\setlength{\tabcolsep}{2.8pt}
\caption{\textbf{Factors at a constant state} (1.4B, 16 heads; state of 110.6M entries per SDM layer in
all rows). Columns as in Table~\ref{tab:mf_heads}; Loss: pretraining loss averaged over the last 5k
steps. Reference row: four factors.}
\label{tab:mf_factors}
\smallskip
\begin{tabular}{@{}rlrr|c|ccc|cc|c@{}}
\toprule
 & & & & & \multicolumn{3}{c|}{PPL, context $T$ $\downarrow$} & \multicolumn{2}{c|}{SLPL, $T$ $\downarrow$} & \\
$H$ & Factor dims & $N$ & Key cols & Loss $\downarrow$ & 32k & 128k & 1M & 32k & 128k & NIAH $\uparrow$ \\
\midrule
16 & $[240,240]$ & 57{,}600 & 7{,}680 & \textbf{2.062}\subg{-0.017} & \textbf{2.021}\subg{-0.016} & \textbf{1.978}\subg{-0.012} & \textbf{1.992}\subg{-0.019} & 3.52\subr{+0.05} & 5.17\subr{+0.26} & \textbf{45.2}\subg{+4.3} \\
16 & $[15,16,16,15]$ & 57{,}600 & 992 & 2.079 & 2.037 & 1.990 & 2.011 & \textbf{3.47} & \textbf{4.91} & 40.9 \\
16 & $[5,5,4,4,4,4,3,3]$ & 57{,}600 & 512 & 2.104\subr{+0.025} & 2.061\subr{+0.024} & 2.011\subr{+0.021} & 2.017\subr{+0.006} & 3.64\subr{+0.17} & 5.23\subr{+0.32} & 41.0\subg{+0.1} \\
\bottomrule
\end{tabular}
\end{table}

\begin{figure}[!htbp]
\centering
\includegraphics[width=\linewidth]{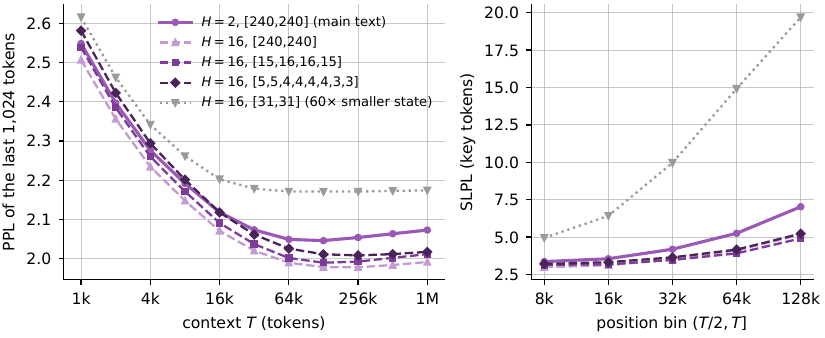}
\caption{\textbf{The five 1.4B runs of Tables~\ref{tab:mf_heads}--\ref{tab:mf_factors}.} Left: PPL of
the last 1{,}024 tokens of 1M-token windows given $T$ tokens of context (the 1M point is the whole
window, starting with BOS). Right: SLPL in the position bin $(T/2,T]$. Solid: the configuration of the
main text, in the SDM colour of the main figures. Dashed: the 16-head runs with 57{,}600 slots per head,
from two (light) to eight (dark) factors. Dotted: the 16-head two-factor run with a 60$\times$ smaller state.}
\label{fig:mf_curves}
\end{figure}
\FloatBarrier

\paragraph{A separate sweep of heads and factors.}
Tables~\ref{tab:mf_sweep} and~\ref{tab:mf_shared} come from a separate sweep with its own weight initialization and training code, so their values compare only within each table.
A separate 1.4B sweep varies the number of heads at a constant state of 57{,}600 slots per head, with
three factors for four to eight heads (two factorizations at five heads) and four factors for 16 heads
(Table~\ref{tab:mf_sweep}); the number of key columns varies between 464 and 992.
From two
heads with two factors to 16 heads with four factors, SLPL decreases from 4.67 to 3.51 and RULER NIAH
increases from 23.0 to 45.6 (36.3 to 40.9 for the same two configurations trained like the main text, Table~\ref{tab:mf_heads}). With two factors, four heads leave room for only 14{,}400 slots per head, and SLPL improves
much less (4.53). 

\begin{table}[!htbp]
\centering\footnotesize
\setlength{\tabcolsep}{2.8pt}
\caption{\textbf{A separate sweep of heads and factors} (1.4B; SLPL in the position bin of about
$[23170, 32768)$; RULER NIAH: exact match (\%) at 32k averaged over the 8 NIAH tasks).
Columns as in Table~\ref{tab:mf_heads}. Loss: pretraining loss averaged over the last 5k steps, as in Table~\ref{tab:mf_factors}. Reference row: the sweep's two-head run.
$^b$Two factors with $N\,{=}\,230{,}400/H^2$, the sweep's rule for two-factor runs.}
\label{tab:mf_sweep}
\smallskip
\begin{tabular}{@{}rlrrr|c|c|c@{}}
\toprule
$H$ & Factor dims & $N$ & Key cols & State & Loss $\downarrow$ & SLPL $\downarrow$ & RULER NIAH $\uparrow$ \\
\midrule
2 & $[240,240]$ & 57{,}600 & 960 & 110.6M & 2.090 & 4.67 & 23.0 \\
4 & $[40,40,36]$ & 57{,}600 & 464 & 110.6M & 2.102\subr{+0.012} & 4.05\subg{-0.62} & 30.6\subg{+7.6} \\
5 & $[30,30,64]$ & 57{,}600 & 620 & 110.6M & 2.079\subg{-0.011} & 3.76\subg{-0.91} & 30.9\subg{+7.9} \\
5 & $[18,20,160]$ & 57{,}600 & 990 & 110.6M & 2.083\subg{-0.007} & 3.89\subg{-0.78} & 28.8\subg{+5.8} \\
8 & $[40,40,36]$ & 57{,}600 & 928 & 110.6M & \textbf{2.077}\subg{-0.013} & 3.72\subg{-0.95} & 32.9\subg{+9.9} \\
16 & $[15,16,16,15]$ & 57{,}600 & 992 & 110.6M & 2.081\subg{-0.009} & \textbf{3.51}\subg{-1.16} & \textbf{45.6}\subg{+22.6} \\
\midrule
4 & $[120,120]^b$ & 14{,}400 & 960 & 27.6M & 2.096\subr{+0.006} & 4.53\subg{-0.14} & 28.6\subg{+5.6} \\
\bottomrule
\end{tabular}
\end{table}
\FloatBarrier

\paragraph{Sharing factors across groups of heads.}
Table~\ref{tab:mf_shared} compares the two-head $[240,240]$ run of Table~\ref{tab:mf_sweep} with three
four-head runs that share one factor within each of $G=2$ pairs of heads. At the same state and nearly
the same key width as the reference (968 vs 960 columns), sharing a 4-entry factor within pairs of heads,
$[120,120\,|\,4]$, gives four heads, and SLPL goes from 4.54 to 4.09 (6.38 vs 7.70 at 128k) and RULER NIAH from 23.0 to 35.6.
At the same state, $[160\,|\,360]$ uses 1{,}360 key columns (SLPL 3.92, RULER NIAH 33.4); $[120\,|\,240]$
keeps 960 columns with half as many slots per head (SLPL 4.47, RULER NIAH 27.0).

\begin{table}[!htbp]
\centering\footnotesize
\setlength{\tabcolsep}{2.0pt}
\caption{\textbf{Sharing factors across groups of heads} (1.4B, after long-context fine-tuning). Factor
dims: per-head factors $|$ factor shared within each of the $G=2$ pairs of heads. Key cols and State as in Table~\ref{tab:mf_heads}; Loss as in Table~\ref{tab:mf_factors}. SLPL as in the Metrics paragraph, at $T=32$k and 128k; RULER NIAH from the sweep's evaluation at
32k, as in Table~\ref{tab:mf_sweep}. Subscripts: difference to the two-head run.}
\label{tab:mf_shared}
\smallskip
\begin{tabular}{@{}rlrrr|c|cc|c@{}}
\toprule
$H$ & Factor dims & $N$ & Key cols & State & Loss $\downarrow$ & SLPL\,32k\,$\downarrow$ & SLPL\,128k\,$\downarrow$ & RULER NIAH $\uparrow$ \\
\midrule
2 & $[240,240]$ & 57{,}600 & 960 & 110.6M & 2.090 & 4.54 & 7.70 & 23.0 \\
4 & $[160\,|\,360]$ & 57{,}600 & 1{,}360 & 110.6M & \textbf{2.074}\subg{-0.016} & \textbf{3.92}\subg{-0.62} & \textbf{6.04}\subg{-1.66} & 33.4\subg{+10.4} \\
4 & $[120\,|\,240]$ & 28{,}800 & 960 & 55.3M & 2.092\subr{+0.002} & 4.47\subg{-0.07} & 7.26\subg{-0.44} & 27.0\subg{+4.0} \\
4 & $[120,120\,|\,4]$ & 57{,}600 & 968 & 110.6M & 2.085\subg{-0.005} & 4.09\subg{-0.45} & 6.38\subg{-1.32} & \textbf{35.6}\subg{+12.6} \\
\bottomrule
\end{tabular}
\end{table}
\FloatBarrier

\paragraph{Cost of selection.}
At a fixed total value width, the per-token FLOPs of the memory reads and writes,
$H(W{+}R)d_{\text{v}}$, do not depend on $H$. However, the top-k selection and the sparse intra-chunk interaction matrices necessary for the efficient parallel algorithm
(Appendix~\ref{sec:efficient-training}) are computed per head, and thus kernel time does depend on the number of heads. Table~\ref{tab:mf_cost} times one SDM layer of the 1.4B model with our SDM kernels, whose product-key top-k takes any number of factors: forward and backward of a training
micro-batch of $4\times8{,}192$ tokens in bf16 with chunk size 128, replayed as one CUDA graph on one NVIDIA
GB300 (a different GPU from the H100 of Figure~\ref{fig:efficiency}). At a constant state, the layer time
grows from 28.9\,ms with two heads and two factors to 49.3\,ms with 16 heads and four factors and 59.5\,ms
with 16 heads and eight factors, and the share of the top-k selection from 1.7\% to 12.3\% and 28.1\%. In the
constant-state rows, the state kernel takes 72--83\% of the layer time with up to four factors, and still 59\%
with eight. Eight factors
need about half the key columns of four (512 against 992), but their selection takes 2.8 times as long
(16.7 against 6.0\,ms). With 16 heads, four factors take as long as two (49.3 against 49.7\,ms): their
selection is slower (6.0 against 3.6\,ms), but their key projections are $7.7\times$ narrower, which shows
in the remaining time (7.9 against 10.3\,ms). The row marked $^\dagger$ is the dense limit: with one factor
of 64 slots per head, every slot selected ($W=R=N=64$) and $\ell_2$-normalized write and read weights in
place of the softmax, the write and read of SDM reduce to the gated delta rule of GDN, and on the same
inputs our kernels match a float64 gated-delta-rule recurrence to rounding. Our kernels are optimized for sparse
updates; in this limit every token writes every slot, and the state kernel dominates (66.9 of the 79.4\,ms
of the layer at chunk size 128, and 39.0 of 51.6\,ms at chunk size 64). 

\begin{table}[!htbp]
\centering\footnotesize
\setlength{\tabcolsep}{2.8pt}
\caption{\textbf{Time of one SDM layer} (1.4B, $W=R=64$, chunk size 128; forward and backward of
$4\times8{,}192$ tokens in bf16, replayed as a CUDA graph on one GB300; median of three processes of 20
steps, which differ by at most 0.3\%). Selection: read and write top-k; State kernel: sparse write and read;
Other: projections, gates, normalization and output; each part is timed alone on the inputs captured from
the layer, and Other is the remainder. Key cols: columns of each of the key and query projections; Key
params: parameters of these two projections together. With chunk size 64 instead of
128, the layer time changes by less than 2\% for up to eight heads. $^\dagger$Dense limit, not at the
constant state of the other rows: one factor of 64 slots per head, all written and read ($W=R=N=64$) with
$\ell_2$-normalized weights instead of the softmax; its state, 122{,}880 entries per layer, is that of the
1.4B GDN (Table~\ref{tab:ladder}) and $900\times$ smaller than that of the constant-state rows.}
\label{tab:mf_cost}
\smallskip
\begin{tabular}{@{}rlrrr|cccc|c@{}}
\toprule
 & & & & & \multicolumn{4}{c|}{Time (ms)} & Selection \\
\cmidrule(lr){6-9}
$H$ & Factor dims & $N$ & Key cols & Key params & Layer & Selection & State kernel & Other & share \\
\midrule
2 & $[240,240]$ & 57{,}600 & 960 & 3.7M & 28.9 & \phantom{0}0.48 & 24.1 & \phantom{0}4.3 & \phantom{0}1.7\% \\
5 & $[30,30,64]$ & 57{,}600 & 620 & 2.4M & 36.6 & \phantom{0}2.89 & 27.6 & \phantom{0}6.0 & \phantom{0}7.9\% \\
8 & $[40,40,36]$ & 57{,}600 & 928 & 3.6M & 40.3 & \phantom{0}4.73 & 29.7 & \phantom{0}5.8 & 11.7\% \\
16 & $[15,16,16,15]$ & 57{,}600 & 992 & 3.8M & 49.3 & \phantom{0}6.04 & 35.3 & \phantom{0}7.9 & 12.3\% \\
16 & $[5,5,4,4,4,4,3,3]$ & 57{,}600 & 512 & 2.0M & 59.5 & 16.73 & 35.0 & \phantom{0}7.8 & 28.1\% \\
16 & $[240,240]$ & 57{,}600 & 7{,}680 & 29.5M & 49.7 & \phantom{0}3.60 & 35.8 & 10.3 & \phantom{0}7.2\% \\
\midrule
16 & $[31,31]$ & 961 & 992 & 3.8M & 39.4 & \phantom{0}1.85 & 30.6 & \phantom{0}6.9 & \phantom{0}4.7\% \\
\midrule
16 & $[64]^\dagger$ & 64 & 1{,}024 & 3.9M & 79.4 & \phantom{0}5.50 & 66.9 & \phantom{0}7.0 & \phantom{0}6.9\% \\
\bottomrule
\end{tabular}
\end{table}
\FloatBarrier

Overall, going from two heads with two factors to 16 heads with four factors makes an SDM layer slower but lowers PPL and SLPL and raises RULER NIAH (Tables~\ref{tab:mf_heads} and~\ref{tab:mf_sweep}).
\fi

\section{Benchmark Stability and the BoolQ Result}
\label{sec:app_boolq}

In Table~\ref{tab:l08_accuracy}, BoolQ is the short-context task whose SDM--GDN delta
changes most between scales ($-0.25$ points at 1.4B, $-7.85$ at 8B). This may partly reflect
benchmark-specific behavior: BoolQ is the least stable task in our evaluation suite.

We use the 24 runs of the scaling ladder (FullAttn, GDN and SDM at the levels of
Table~\ref{tab:ladder}) and their evaluations during pretraining, with the scoring metrics of Table~\ref{tab:l08_accuracy}. 
The models reported in Table~\ref{tab:l08_accuracy} have additionally undergone long-context fine-tuning.
For each run we compute the fraction of its consecutive evaluations whose score is strictly higher than the
previous one and average this fraction over runs; a run is counted as monotone if every one of its
evaluations improves on the previous one. Both describe how consistently a task's score increases during
pretraining.

\begin{table}[!htb]
\centering\small
\caption{\textbf{Evaluation-to-evaluation stability across the 24 ladder runs} (pretraining
evaluations). Evaluations improving: fraction of a run's consecutive evaluations that improve on the previous
one, averaged over runs. A run is counted as monotone if every one of its successive evaluations improves on
the previous one. BoolQ improves least often; it and CommonsenseQA are monotone in no run.}
\label{tab:boolq-stability}
\smallskip
\begin{tabular}{@{}lcc@{}}
\toprule
Benchmark & Evaluations improving (\%) & Monotone runs \\
\midrule
HellaSwag        & 100 & 24/24 \\
TQA              & \mysp96 & 21/24 \\
RACE.mid         & \mysp88 & 15/24 \\
PIQA             & \mysp86 & 14/24 \\
RACE.high        & \mysp86 & 14/24 \\
NaturalQuestions & \mysp86 & 14/24 \\
ARC-easy         & \mysp80 & 10/24 \\
ARC-challenge    & \mysp71 & \mysp8/24 \\
WinoGrande       & \mysp71 & \mysp8/24 \\
OpenBookQA       & \mysp65 & \mysp2/24 \\
MMLU             & \mysp60 & \mysp5/24 \\
HumanEval+       & \mysp59 & \mysp4/24 \\
GSM8K            & \mysp56 & \mysp3/24 \\
CommonsenseQA    & \mysp49 & \mysp0/24 \\
BoolQ            & \mysp40 & \mysp0/24 \\
\bottomrule
\end{tabular}
\end{table}

HellaSwag improves at every evaluation of every run, and TQA, RACE, NaturalQuestions and PIQA are close
behind. BoolQ improves least often (40\% of consecutive evaluations on average) and, like CommonsenseQA,
is monotone in none of the 24 runs, so we treat the 8B BoolQ gap with caution.

\ifmultifactor\else\newpage\fi 

\FloatBarrier 
\section{RULER Per-Task Accuracy by Sequence Length}
\label{sec:ruler-per-task}

\begin{figure}[htbp]
    \centering
    \includegraphics[width=\textwidth]{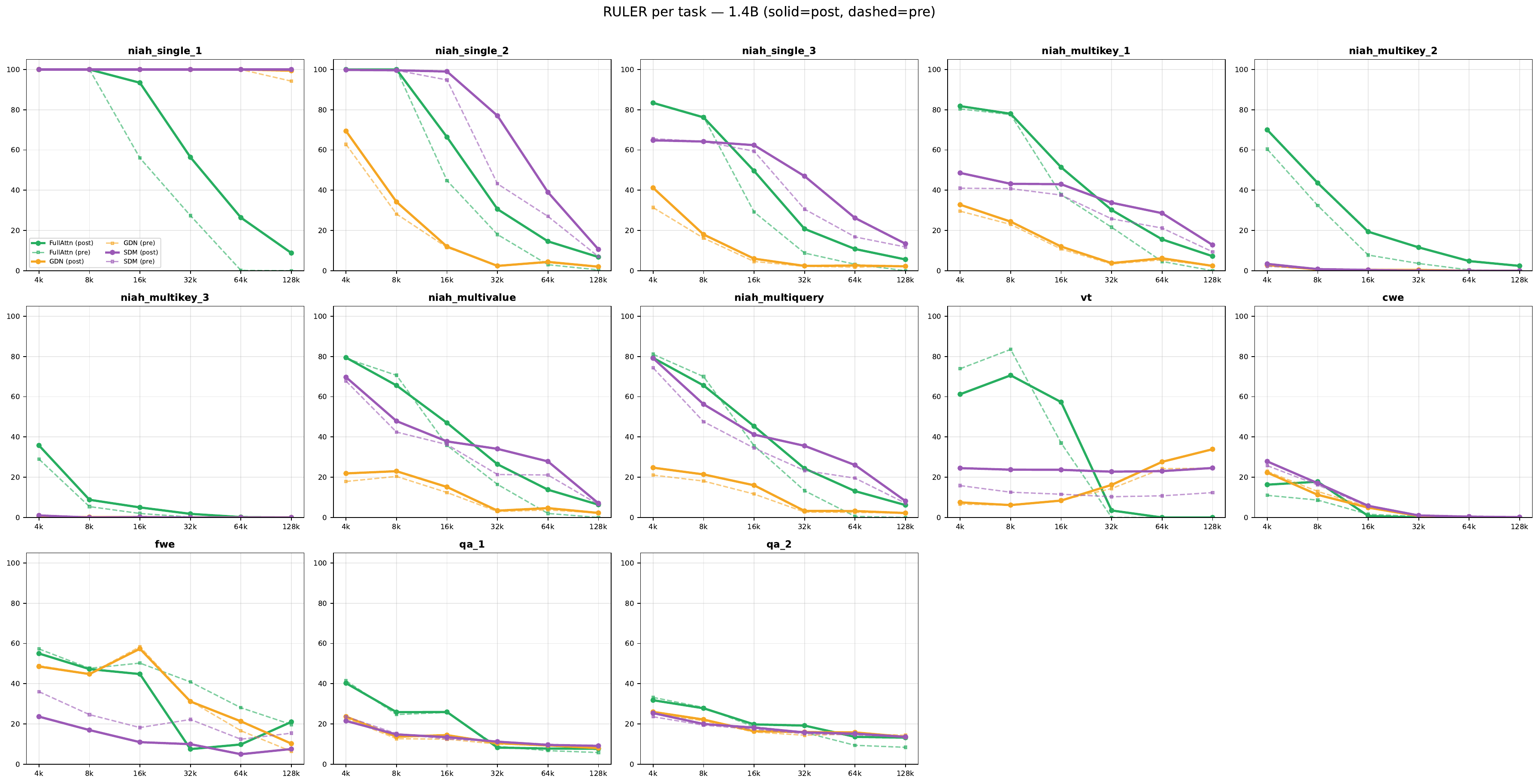}
    \caption{\textbf{RULER per-task accuracy (\%) by sequence length at 1.4B scale.} Solid lines: after 128k post-training. Dashed lines: pre-trained only. SDM maintains strong recall on NIAH single-needle tasks across all lengths, while Full Attention degrades rapidly beyond its training length. GDN struggles on multi-step retrieval (single\_2, single\_3, multiquery). Surprisingly, on the variable tracking task, GDN is better on longer sequence lengths. It may be due to the design of this task: it clusters distractor variables early in the sequence, reducing their influence for long sequences.
    \label{fig:ruler_per_task_1.4b}}
\end{figure}

\begin{figure}[htbp]
    \centering
    \includegraphics[width=\textwidth]{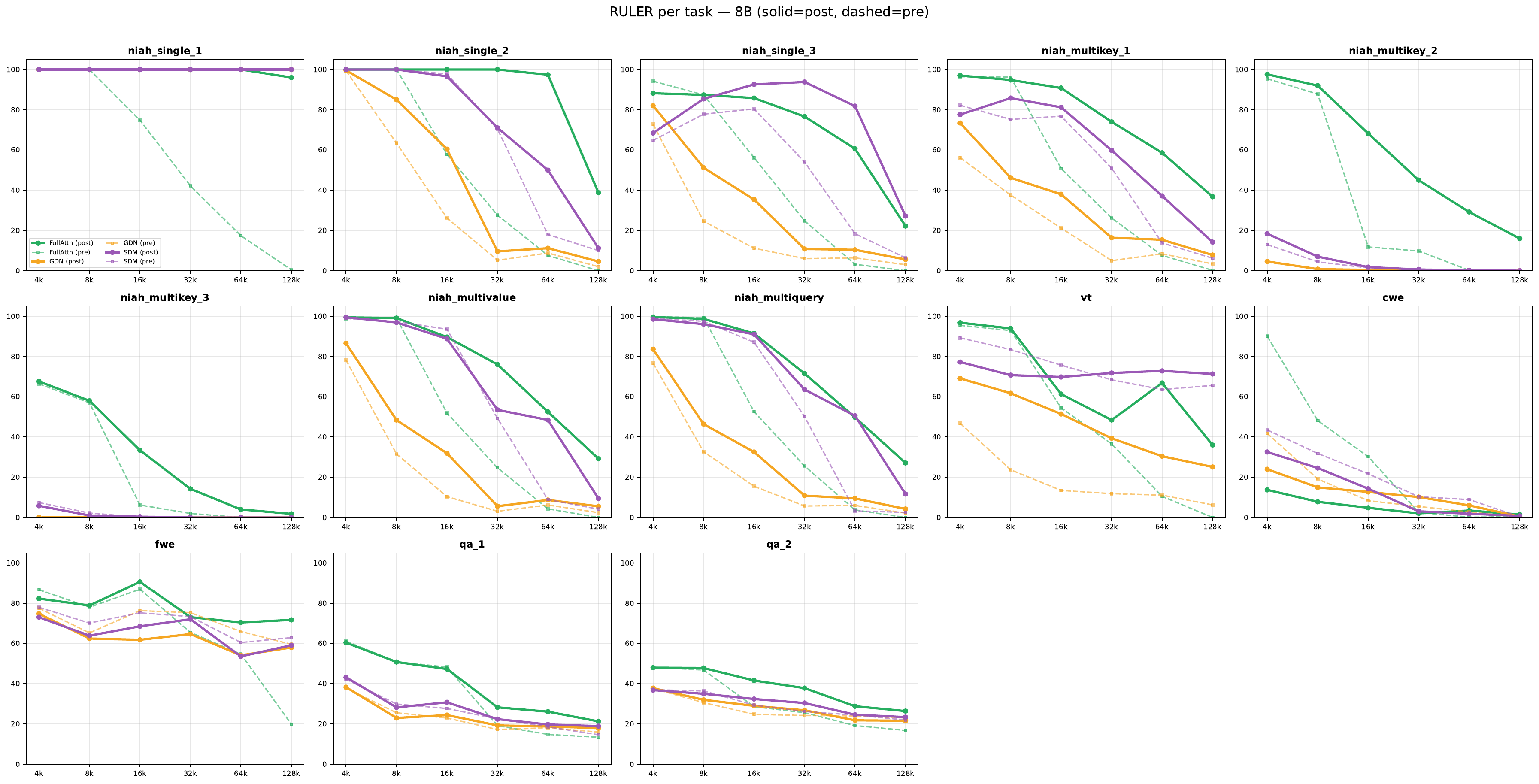}
    \caption{\textbf{RULER per-task accuracy (\%) by sequence length at 8B scale.} Solid lines: after 128k post-training. Dashed lines: pre-trained only. At 8B, post-training dramatically improves FullAttn (green), which achieves near-perfect recall on single-needle tasks. SDM maintains a strong recall and outperforms FullAttn on the single\_3 and vt tasks. The multikey\_2 task is challenging for all fixed-state models.
    \label{fig:ruler_per_task_8b}}
\end{figure}

Figure~\ref{fig:ruler_per_task_1.4b} and Figure~\ref{fig:ruler_per_task_8b} show the per-task RULER accuracy at each sequence length for different architectures at 1.4B scale and 8B scale respectively. All 13 RULER tasks are shown. Solid lines denote post-trained models (128k sequence length finetuned); dashed lines denote pre-trained models.

\newpage
\section{Hybridizing SDM with FullAttn}
\label{sec:app_mixed_hybrid}

We explore a three-way hybrid architecture combining SWA (local), FullAttn (global), and SDM (global) layers within the same model.
In this configuration, global layer positions alternate between FullAttn and SDM:
for 5 global layers, the pattern is FA, SDM, FA, SDM, FA.
All other layers remain SWA with a window of 128 tokens.
We denote this architecture SWA:(FA/SDM).

\begin{figure}[h]
    \centering
    \includegraphics[width=0.7\textwidth]{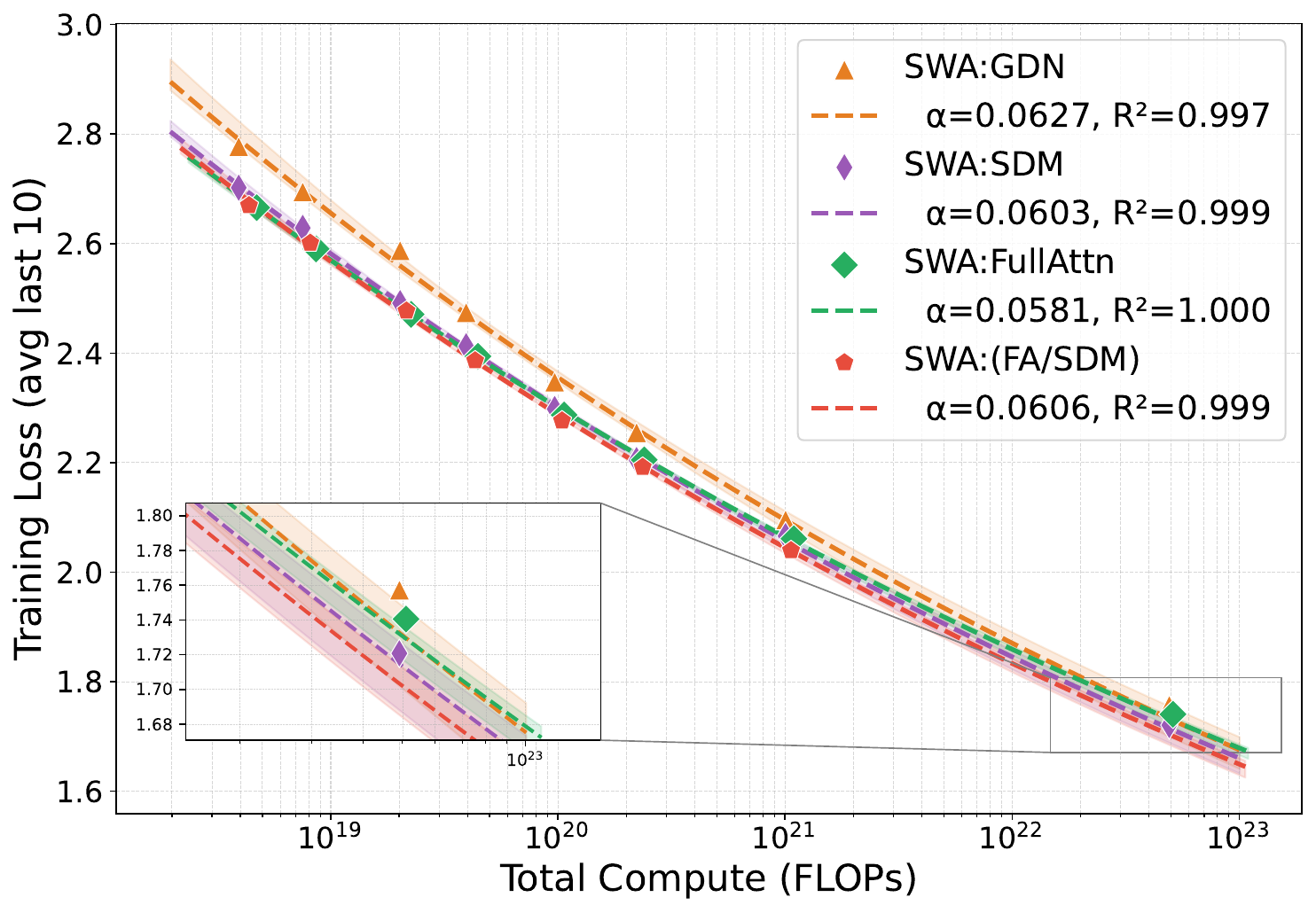}
    \caption{Training loss scaling law for four architectures at levels 1--8.}
    \label{fig:scaling_mixed}
\end{figure}

As shown in Figure~\ref{fig:scaling_mixed},  the 3-way hybrid architecture outperforms both 2-way hybrids at larger scale thanks to its better scaling coefficient. 
This hints at the fact that SDM and FullAttn might provide complementary capabilities and also at the possibility of augmenting existing local-global architectures with SDM layers.

\begin{table}[h]
\centering
\footnotesize
\begin{tabular}{lcccc}
\toprule
 & SWA:FA & SWA:(FA/SDM) & SWA:SDM & SWA:GDN \\
\midrule
Validation text NLL $\downarrow$ & 2.658 & \textbf{2.645} & 2.654 & 2.684 \\
Validation code NLL $\downarrow$ & 0.798 & \textbf{0.797} & 0.821 & 0.849 \\
Avg. Accuracy (\%) $\uparrow$ & 38.1 & \textbf{38.6} & \textbf{38.6} & 37.9 \\
RULER (6-task, 4k--8k) $\uparrow$ & \textbf{80.1} & 64.3 & 56.9 & 32.9 \\
\bottomrule
\end{tabular}
\caption{Performance of the mixed SWA:(FA/SDM) hybrid at 1.4B scale (L08, pre-trained). RULER is averaged across 6 tasks (niah\_single 1/2/3, multikey\_2, multiquery, vt) at sequence lengths 4k--8k (within training length).}
\label{tab:mixed_hybrid}
\end{table}

\section{Memory Utilization}
\label{sec:mem-utilization}

\subsection{During Training}
\begin{figure}[htbp]
    \centering
    \includegraphics[width=\textwidth]{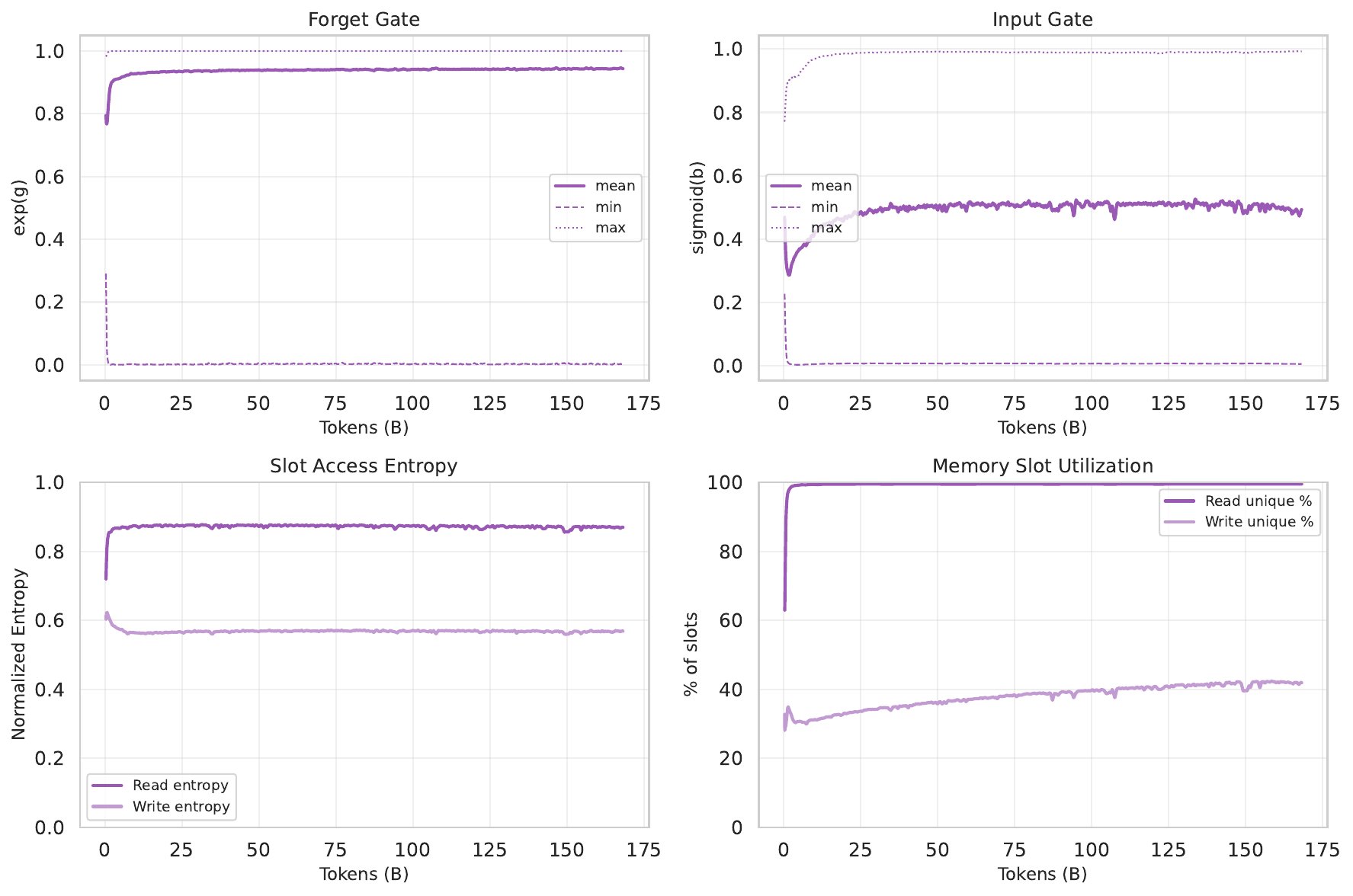}
    \caption{\textbf{SDM memory utilization during training (1.4B, L08).} Statistics are accumulated over 10 training steps ($\sim$21M tokens) per data point and averaged across 5 SDM layers. \textit{Top}: forget and input gate statistics (min/mean/max). \textit{Bottom left}: normalized entropy of slot access distributions. \textit{Bottom right}: percentage of unique memory slots accessed.}
    \label{fig:mem_utilization}
\end{figure}

Figure~\ref{fig:mem_utilization} shows the evolution of key memory access statistics during training of the SDM model at 1.4B scale (L08). All metrics are averaged across the 5 SDM layers.

The forget gate (top left) converges early to $\alpha_t \approx 0.95$. The minimum forget gate stays near zero, indicating that some slots are fully decayed when overwritten. The input gate (top right) stabilizes at $\beta_t \approx 0.50$, with the maximum and minimum reaching $\sim$1 and $\sim$0 respectively. Overall, the model learns a wide dynamic range for both forget and input strength, showing the importance of both mechanisms in the model.

The read utilization (bottom right) quickly reaches 100\% of the memory, showing that no memory slot is left unutilized. 
The write utilization (bottom right) grows steadily from 28\% to 42\% during training, showing that the model learns to diversify its write patterns over time, while still keeping some part of the memory not modified by in-context updates.

\subsection{Access Patterns}
\label{app:ablations}

The values are collected by an offline probe run on each ablation model's final checkpoint using a fixed held-out batch from coding data. Concretely, for every run, the script reads the first $B=8$ documents whose tokenized length is at least $T=8192$, truncates each document to 8192 tokens, and performs a single forward pass on this $8\times 8192$ batch. During this pass, forward hooks are attached to every SDM layer to capture the post-softmax weights over the selected top-$k$ memory slots for each token and attention head, with $k=R$ on the read side and $k=W$ on the write side. All reported statistics are then computed from these per-token top-$k$ distributions and aggregated across tokens, heads, and layers.

\paragraph{Cumulative Mass in top-k Keys.}
\label{app:cum_mass}
For the cumulative mass in the top-$m$ keys, let $p\in\mathbb{R}^k$ denote the post-softmax distribution over the selected keys for a single token, and let $p_{(1)}\geq p_{(2)}\geq\cdots\geq p_{(k)}$ be the same values sorted in descending order. We then define the cumulative mass curve as
$$
C(m)=\sum_{i=1}^{m} p_{(i)}, \qquad m=1,\dots,k.
$$
This quantity measures how quickly the probability mass concentrates in the highest-weighted keys: if $C(m)$ rises rapidly toward $1$, then only a few keys account for most of the mass and the distribution is highly peaked; if it rises more slowly, the controller is making broader use of its available top-$k$ budget. 

\paragraph{Effective Number of Keys per Token.}
\label{app:eff_keys}
For the effective-keys calculation, we compute the entropy of the per-token top-$k$ distribution,
$$
\mathrm{ent}(p)=-\sum_{i=1}^{k} p_i \log p_i,
$$
and convert it into an entropy-based effective number of keys,
$$
\mathrm{eff\_keys}(p)=\exp(\mathrm{ent}(p)).
$$
This is equal to $1$ for a delta-like distribution concentrated on a single key and approaches $k$ when the distribution is close to uniform over the selected budget. To make the quantity directly comparable across runs with different values of $k$, we normalize it as
$$
u(p)=\frac{\mathrm{eff\_keys}(p)}{k}\in(0,1].
$$
Thus, $u(p)\approx 1$ indicates near-uniform use of the available top-$k$ slots, whereas $u(p)\ll 1$ indicates that the controller effectively relies on only a small fraction of them.

\paragraph{Performance across different read and write key pairs for SDM.}
All values are reported at the 1.4B model scale after pre-training at an 8k sequence length.

\begin{table}[t]
\centering
\footnotesize
\setlength{\tabcolsep}{5pt}
\begin{tabular}{lcccccc}
\toprule
Run & Train text NLL $\downarrow$ & $\Delta$ & Avg. RULER $\uparrow$ & $\Delta$ & Avg. Reasoning $\uparrow$ & $\Delta$ \\
\midrule
W64\_R128 & \textbf{2.6703} & -0.0016 & 0.4011 & -0.0164 & 0.3842 & -0.0026 \\
W64\_R64 & 2.6719 & \phantom{+}0.0000 & \textbf{0.4176} & \phantom{+}0.0000 & 0.3868 & \phantom{+}0.0000 \\
W32\_R64 & 2.6726 & +0.0007 & 0.3501 & -0.0675 & \textbf{0.3890} & +0.0022 \\
SWA:FullAttn & 2.6729 & +0.0010 & 0.3474 & -0.0702 & 0.3805 & -0.0063 \\
W32\_R128 & 2.6737 & +0.0018 & 0.4039 & -0.0136 & 0.3821 & -0.0047 \\
W64\_R32 & 2.6746 & +0.0027 & 0.3764 & -0.0412 & 0.3887 & +0.0019 \\
W32\_R32 & 2.6766 & +0.0047 & 0.3612 & -0.0564 & 0.3871 & +0.0003 \\
\bottomrule
\end{tabular}
\smallskip
\caption{\textbf{Comparison across varying reads (R) and writes (W)}. Deltas are computed w.r.t. \texttt{W64\_R64}. Train text NLL: the text dataset training loss at the final training step (lower is better); higher is better for Avg. RULER and Avg. Reasoning. Bold indicates the best value in each non-delta metric column. RULER averages 6 sub-tasks at lengths 4k--128k, for computational reasons: niah\_single 1/2/3, multikey 2, multiquery and vt.
}
\label{tab:ablation-toprw-base-runs-filtered}
\end{table}

\section{Memory Addressing: Collisions and Key Structure}
\label{sec:app_collisions}

Every token writes to $W=64$ slots per head, so two tokens can interact through the slots
they share. We quantify this overlap and then
isolate it as one of the causes of SDM's weakness (compared to FullAttn) on dense multi-key retrieval. All
measurements use the 8B model.

\subsection{Write-Set Overlap}

We define the collision between two tokens as the Jaccard overlap of their write sets: $0$
for disjoint slots, $1$ for identical ones. Averaged over token pairs on ordinary text, the mean
Jaccard overlap is ${\approx}0.24$, against ${\approx}0$ for random
writing. Addressing is therefore content-dependent as intended; the overlap is similar for nearby and
distant token pairs.

Heads specialize. At one extreme a head writes near-disjointly (${\approx}3\%$ overlap,
${\approx}30$k distinct slots touched over an 8k-token sequence); at the other a head
behaves as a working memory, concentrating on ${\approx}1.2$k slots with ${\approx}43\%$
overlap and refreshing them in place.

Collisions are also useful: they are how SDM overwrites stale content. In an update test, in
which an entity's value is changed several times in context, successive writes land on the
same slots $130$--$350\times$ above chance, and the model returns the most recent value in
47 of 48 trials.

\subsection{What Limits Dense Multi-Key Recall}
\label{sec:app_keytype}

Table~\ref{tab:l08_accuracy} shows that SDM, like other fixed-state models, remains far
behind full attention on multikey 2 and multikey 3. This is partially addressable through training: in a dedicated copy-training run, SDM learns to store and reproduce a full
16k-token text, far more information than the few hundred key--value pairs these tasks
require. To separate the remaining causes we run a RULER-style haystack of key--value pairs
and vary only the \emph{type} of key, holding the values fixed, at 200 pairs (the real
RULER multi-key task uses word keys, number values and 426 pairs, so this probe is
representative of it). Exact-match accuracy:

\begin{table}[t]
\centering\small
\caption{\textbf{Exact-match accuracy (\%) by key type}, 8B, 200 key--value pairs. Only the
key type varies. Down a column, key structure alone moves SDM roughly tenfold; across a
row, accuracy orders by state size at every key type.}
\label{tab:keytype}
\smallskip
\begin{tabular}{@{}lccc@{}}
\toprule
Key type & SDM & GDN & FullAttn \\
\midrule
distinct letters & 30 & \mysp2 & 95 \\
distinct words   & 19 & \mysp8 & 90 \\
numbers          & \mysp7 & \mysp6 & 99 \\
UUIDs            & \mysp3 & \mysp3 & 99 \\
\bottomrule
\end{tabular}
\end{table}

Two effects are visible, along the two axes of the table. Reading down a column, at fixed
state size, key structure alone moves SDM roughly tenfold, from 3\% on UUIDs to 30\% on
single letters: similar keys collide in the product-key address space, and this part of the
gap is a learned allocation behaviour --- distinguishing ``overwrite'' from ``store
separately'', since collisions are also the mechanism by which SDM supersedes stale values
--- that better addressing and training can mitigate (Section~\ref{sec:app_mktrain}).
One must keep in mind, however, that while colliding writes may indeed be detrimental when both pieces of information must be remembered independently, collisions are also how SDM, and the delta rule more generally, updates previous facts with newer information. Thus, reducing collisions indiscriminately, e.g.\ with dedicated auxiliary losses, may not improve long-context performance. 
Reading across a row, at fixed keys,
accuracy orders strictly by state size at every key type: full attention, which compresses
nothing, then SDM, then GDN, whose state is too small for the task regardless of
addressing.

These two limitations can and do coexist.
Indeed, even on the most favourable keys, i.e.\ the most distinct ones, SDM reaches 30\% against
95\% for full attention, so improving the addressing within a fixed state will not close the
gap on its own: what remains is the cost of compressing an unbounded context into a bounded
state, the same quantity the memory-size ablation of Table~\ref{tab:sdm_ablations} traces.
Dense multi-key retrieval is therefore an open problem for fixed-state models generally
rather than for SDM specifically; what SDM adds is the freedom to place
similar-but-distinct information in separate slots, which a single dense state does not have.

\subsection{Targeted Multi-Key Training}
\label{sec:app_mktrain}

The key-type probe suggests that part of the multi-key gap is learned addressing behaviour.
To test how much of it training on the task recovers, we fine-tune the three 8B post-trained
models of Table~\ref{tab:l08_accuracy} on RULER-style multi-key retrieval and re-evaluate them
on RULER.

\paragraph{Setup.} The three models are trained identically, and only the global layer differs:
400 steps at an 8k context with a global batch of 4 sequences (${\approx}13$M tokens), using
AdamW with a freshly initialized optimizer, a learning rate of $10^{-5}$ (50 warmup steps, then
constant), $\beta=(0.9, 0.95)$, weight decay 0.1 and gradient clipping at 1.0. The loss is
applied to the answer only. The training mix contains 50\% multikey 2 haystacks, which
reproduce the RULER format exactly: the context consists only of lines of the form
\emph{One of the special magic numbers for $\langle$adjective$\rangle$-$\langle$noun$\rangle$
is: $\langle$7 digits$\rangle$}, roughly 350 key--value pairs per 8k sequence, and one key is
queried. A further 20\% are drawn from the three multi-key tasks (multikey 1, 2 and 3), 12\% are
single-needle retrieval and 18\% is code pre-training data.
Overall, multikey 2 accounts for ${\approx}58\%$ of the mix, multikey 3 for ${\approx}8\%$ and
multikey 1 for ${\approx}4\%$. The generator matches RULER's templates, so results up to 8k measure how well each model fits
the task in distribution. No sequence is longer than 8k tokens, so every result beyond 8k
measures length extrapolation. We evaluate 500 samples per task and length.

\begin{figure}[h]
    \centering
    \includegraphics[width=\textwidth]{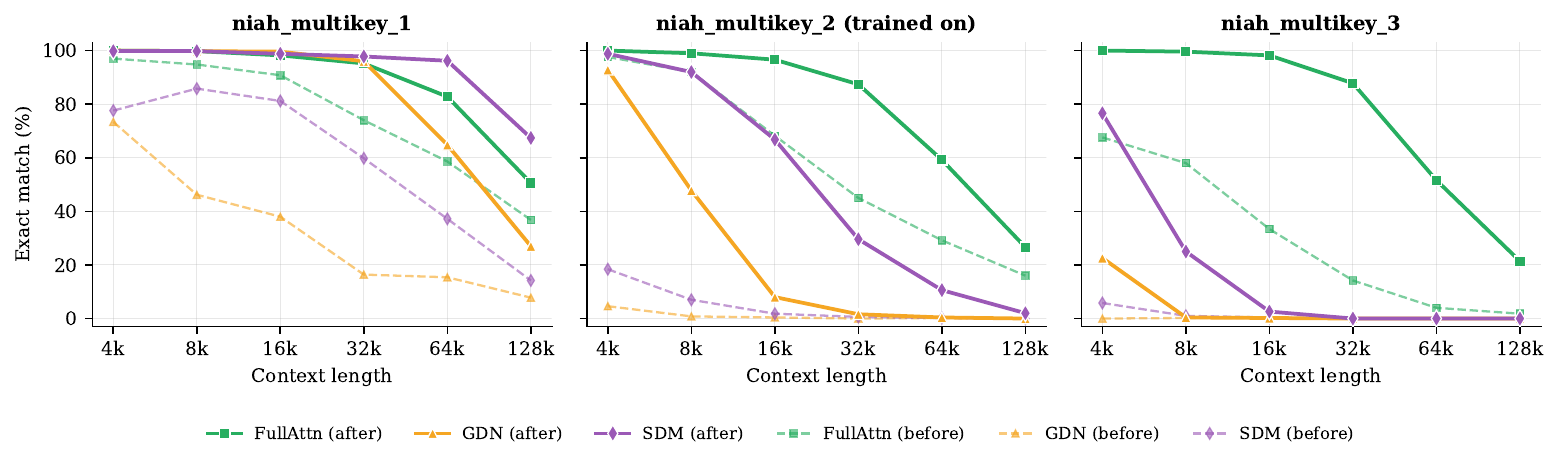}
    \caption{\textbf{RULER multi-key accuracy before and after multi-key training at 8B scale.}
    Dashed lines: the post-trained models of Table~\ref{tab:l08_accuracy}. Solid lines: after
    400 steps of multi-key training at 8k context. All lengths beyond 8k are extrapolation.}
    \label{fig:mk_ft}
\end{figure}

\begin{table}[h]
\centering\small
\caption{\textbf{Multikey 2 exact match (\%) before and after multi-key training}, 8B, 500
samples per length. Avg.\ is the mean over the six lengths, as in Table~\ref{tab:l08_accuracy}.}
\label{tab:mk_ft}
\smallskip
\begin{tabular}{@{}llccccccc@{}}
\toprule
 & & 4k & 8k & 16k & 32k & 64k & 128k & Avg. \\
\midrule
FullAttn                  & before & 97.6 & 92.0 & 68.2 & 45.0 & 29.2 & 16.0 & 58.0 \\
                          & after  & 100.0 & 99.0 & 96.6 & 87.4 & 59.2 & 26.6 & 78.1 \\
\midrule
GDN                       & before & 4.6 & 0.8 & 0.4 & 0.0 & 0.2 & 0.0 & 1.0 \\
                          & after  & 92.8 & 47.8 & 8.0 & 1.6 & 0.4 & 0.0 & 25.1 \\
\midrule
SDM                       & before & 18.4 & 7.0 & 1.8 & 0.6 & 0.2 & 0.0 & 4.7 \\
                          & after  & 98.8 & 92.0 & 66.8 & 29.6 & 10.6 & 2.0 & 50.0 \\
\bottomrule
\end{tabular}
\end{table}

\paragraph{Results.} Training lifts every model on multikey 1 and 2 at the lengths it was
trained on (4k and 8k): on multikey 2 at 4k, SDM rises from 18\% to 99\% and GDN from 5\% to 93\% (Table~\ref{tab:mk_ft}). The two
fixed-state models then extrapolate very differently. SDM retains 67\% at 16k and 30\% at 32k,
against 8\% and 2\% for GDN, which stays below 2\% from 32k on and scores 0 of 500 at 128k
(95\% upper bound 0.7\%). Multikey 1, a small part of the mix, shows the same pattern more
strongly: after training, SDM is the most accurate of the three models beyond 32k, with 96\% at
64k and 67\% at 128k against 83\% and 51\% for full attention (Figure~\ref{fig:mk_ft}).
Multikey 3, whose keys and values are UUIDs, improves at 4k (from 6\% to 77\% for SDM) but
collapses beyond 8k for both fixed-state models. This is consistent with UUID keys being the
hardest case in Table~\ref{tab:keytype}.

Training therefore narrows the gap to full attention without closing it. On multikey 2, full
attention still leads at every length beyond 8k (87\% against 30\% for SDM at 32k), and the
ordering FullAttn $>$ SDM $>$ GDN of Table~\ref{tab:keytype} is preserved. This is consistent
with Section~\ref{sec:app_keytype}. 

\section{Compute Resources}
\label{sec:app_compute_resources}

\ifmultifactor Except for the multi-factor experiments of Appendix~\ref{sec:app_multifactor}, which also used NVIDIA GB300 GPUs, all\else All\fi{} experiments in this work were conducted using NVIDIA H100 Hopper 80GB GPUs using the latest stable PyTorch release available at run time.
The models are implemented in PyTorch~\citep{paszke2019pytorch}; the attention layers use xFormers~\citep{xFormers2022}, and the GDN layers use the kernels of the FLA library~\citep{yang2024fla}.
We estimate the GPU-hour usage of this work for pre-training, post-training, ablations and evaluation to be around 300k GPU hours in total.
In a training-step benchmark at 1.4B scale (2 nodes, four 8k-token sequences per GPU, chunk size $C=64$; Appendix~\ref{sec:efficient-training}), a step takes 1.017\,s for SDM against 0.849\,s for GDN, a 20\% overhead (16\% at 8B).
The ${\sim}54$k-token prefill crossover between SDM and FullAttn in Figure~\ref{fig:efficiency} is linearly interpolated between the measured 32k and 65k contexts.

\section{Licenses of Existing Assets}
\label{sec:app_licenses}

Table~\ref{tab:licenses} lists the licenses and terms of use of the evaluation benchmarks used in this work\ifmultifactor{} and of the model used as an oracle in Appendix~\ref{sec:app_multifactor}\fi, as stated by their creators.

\ifmultifactor\newcommand{\SDMllamarow}{Llama-3.1-8B~\citep{grattafiori2024llama3herdmodels} (SLPL oracle, Appendix~\ref{sec:app_multifactor}) & Llama 3.1 Community License \\}\else\newcommand{\SDMllamarow}{}\fi
\begin{table}[h]
\centering
\small
\caption{\textbf{Licenses of the evaluation benchmarks}\ifmultifactor{} and of the oracle model\fi{} used in this work, as stated by their creators.}
\label{tab:licenses}
\smallskip
\begin{tabular}{@{}ll@{}}
\toprule
Asset & License or terms of use \\
\midrule
HellaSWAG~\citep{zellers2019hellaswagmachinereallyfinish} & MIT \\
WinoGrande~\citep{sakaguchi2019winograndeadversarialwinogradschema} & CC BY (data), Apache 2.0 (code) \\
ARC~\citep{clark2018thinksolvedquestionanswering} & CC BY-SA 4.0 \\
PIQA~\citep{bisk2019piqareasoningphysicalcommonsense} & Academic Free License 3.0 \\
OpenBookQA~\citep{OpenBookQA2018} & Apache 2.0 \\
RACE~\citep{lai2017racelargescalereadingcomprehension} & Non-commercial research purposes only \\
CommonsenseQA~\citep{talmor2019commonsenseqaquestionansweringchallenge} & MIT \\
BoolQ~\citep{clark2019boolqexploringsurprisingdifficulty} & CC BY-SA 3.0 \\
TriviaQA~\citep{joshi-etal-2017-triviaqa} & Apache 2.0 (code and data) \\
HumanEval~\citep{chen2021evaluatinglargelanguagemodels}, HumanEval+~\citep{evalplus} & MIT; Apache 2.0 \\
NaturalQuestions~\citep{kwiatkowski-etal-2019-natural} & CC BY-SA 3.0 \\
MMLU~\citep{hendryckstest2021} & MIT \\
GSM8K~\citep{cobbe2021gsm8k} & MIT \\
RULER~\citep{hsieh2024rulerwhatsrealcontext} & Apache 2.0 \\
\SDMllamarow
\bottomrule
\end{tabular}
\end{table}

\end{document}